\documentclass{article}

\usepackage{iclr2027_conference,times}
\usepackage[T1]{fontenc}
\usepackage[utf8]{inputenc}
\usepackage{microtype}
\usepackage{inconsolata}
\usepackage{graphicx}
\usepackage{booktabs}
\usepackage{amsmath}
\usepackage{amssymb}
\usepackage{tabularx}
\usepackage{array}
\usepackage{makecell}
\usepackage{hyperref}
\usepackage{url}
\usepackage{wrapfig}
\title{FinEvolveBench: A Benchmark for Self-Evolving Agents under Challenging Experience Regimes}

\author{%
\begin{tabular*}{0.95\textwidth}{@{\extracolsep{\fill}}llll@{}}
\rule{0pt}{24pt}Yining Zhu &
Zihao Deng &
Wang Leiming &
Jingfei Lu \\
\rule{0pt}{24pt}Junbo Wang &
Yuke Li &
Jikun Shen &
Chuncheng Ran
\end{tabular*}%
}

\newlength{\FEoriginalheadsep}
\newlength{\FEoriginalfootskip}
\AddToHook{cmd/@outputpage/before}{%
  \ifnum\value{page}>1\relax
    \ifnum\value{page}<11\relax
      \ifnum\value{page}=6\relax
      \else
        \addtolength{\headsep}{-16pt}%
        \addtolength{\footskip}{16pt}%
      \fi
    \fi
  \fi
}

\AddToHook{cmd/@outputpage/after}{%
  \setlength{\headsep}{\FEoriginalheadsep}%
  \setlength{\footskip}{\FEoriginalfootskip}%
}

\begin{document}
\iclrfinaltrue
\maketitle
\lhead{}

% 有关经验机制设计的 LLM Agent 的文章研究很多，但是它们都存在仅在各自声称的场景上有效的问题。目前没有，as far as we know，目前没有文章能够从场景的角度去分析各个经验机制的有效性。本篇文章对目前一些常用的数据集进行了场景类型的总结分类，对现有方法在不同场景的适用性做了总结，并提出了一个新的 FinEvolveBench 数据集，弥补现有数据集在维度上的不足，并用现有不同的经验机制方法在此benchmark上做了实验，发现效果均有下降，体现本场景数据集的必要性。

\begin{abstract}
Large language model (LLM) agents increasingly rely on external experience to continually adapt to changing environments without modifying their underlying models. Recent experience mechanisms have demonstrated promising results across diverse tasks. However, their effectiveness is typically evaluated within individual benchmark settings, and how experience mechanisms generalize across different scenarios remains insufficiently explored. In this work, we present a scenario-oriented analysis of experience mechanisms for LLM agents. We characterize existing evaluation scenarios along four dimensions: outcome observability, credit assignment complexity, environmental dynamics, and experience reusability. Our analysis shows that existing benchmarks often evaluate experience mechanisms under scenarios where at least one dimension is comparatively favorable, leaving more challenging combinations of scenario properties underexplored. To address this gap, we introduce \textsc{FinEvolveBench}, a reproducible benchmark built on a chronological stream of rich financial news and market data that enables systematic evaluation of experience-based self-evolution under challenging experience regimes characterized by noisy feedback, ambiguous credit assignment, environmental non-stationarity, and limited experience reusability. Experiments show that existing approaches exhibit substantially reduced or inconsistent gains in this setting, highlighting the scenario-dependent nature of experience mechanisms and the challenge of maintaining valid experience under changing environments.
\end{abstract}

\section{Introduction}
Large language model (LLM) agents are increasingly expected to solve complex real-world tasks while continually adapting to changing environments, moving beyond static, one-shot settings. A promising direction is \emph{experience-based self-evolution}, where agents improve at test time by accumulating, reflecting on, updating, and reusing experiences without modifying the backbone model. Existing approaches, including reflection-based methods, external memory systems, and experience retrieval mechanisms, have demonstrated encouraging gains across reasoning, software engineering, web interaction, embodied environments, and long-horizon tasks~\cite{shinn2023reflexion,zhao2024expel,wang2023voyager,zheng2025lifelongagentbench}. These developments suggest that external experience can serve as an effective mechanism for continual self-evolution. However, the effectiveness of such mechanisms is typically established within specific benchmark settings, and their applicability across different tasks and environmental conditions remains underexplored.

We argue that experience mechanisms should not be evaluated independently of the scenarios in which experience is acquired and reused. In our review of existing evaluation settings, we characterize these scenarios along four dimensions: \emph{outcome observability}, \emph{credit assignment complexity}, \emph{environmental dynamics}, and \emph{experience reusability}. These dimensions directly shape how an agent should acquire, refine, update, and retain experience. For example, coding environments provide relatively observable outcomes through execution and tests, whereas web and embodied tasks often involve more ambiguous credit assignment, calling for different strategies for experience refinement and update. Likewise, open-world and social environments are typically more non-stationary, where previously useful experience may become outdated and thus require explicit revision or forgetting; in contrast, experience can remain valid for much longer in environments governed by stable rules, such as reproducible simulators or static reasoning tasks. These observations motivate a scenario-aware evaluation of experience mechanisms: their effectiveness should be interpreted jointly with the scenario properties under which they are learned and deployed, rather than inferred from performance in a single setting.

\begin{figure}[h]
    \vspace{-10pt}
    \centering
    \includegraphics[width=0.8\textwidth]{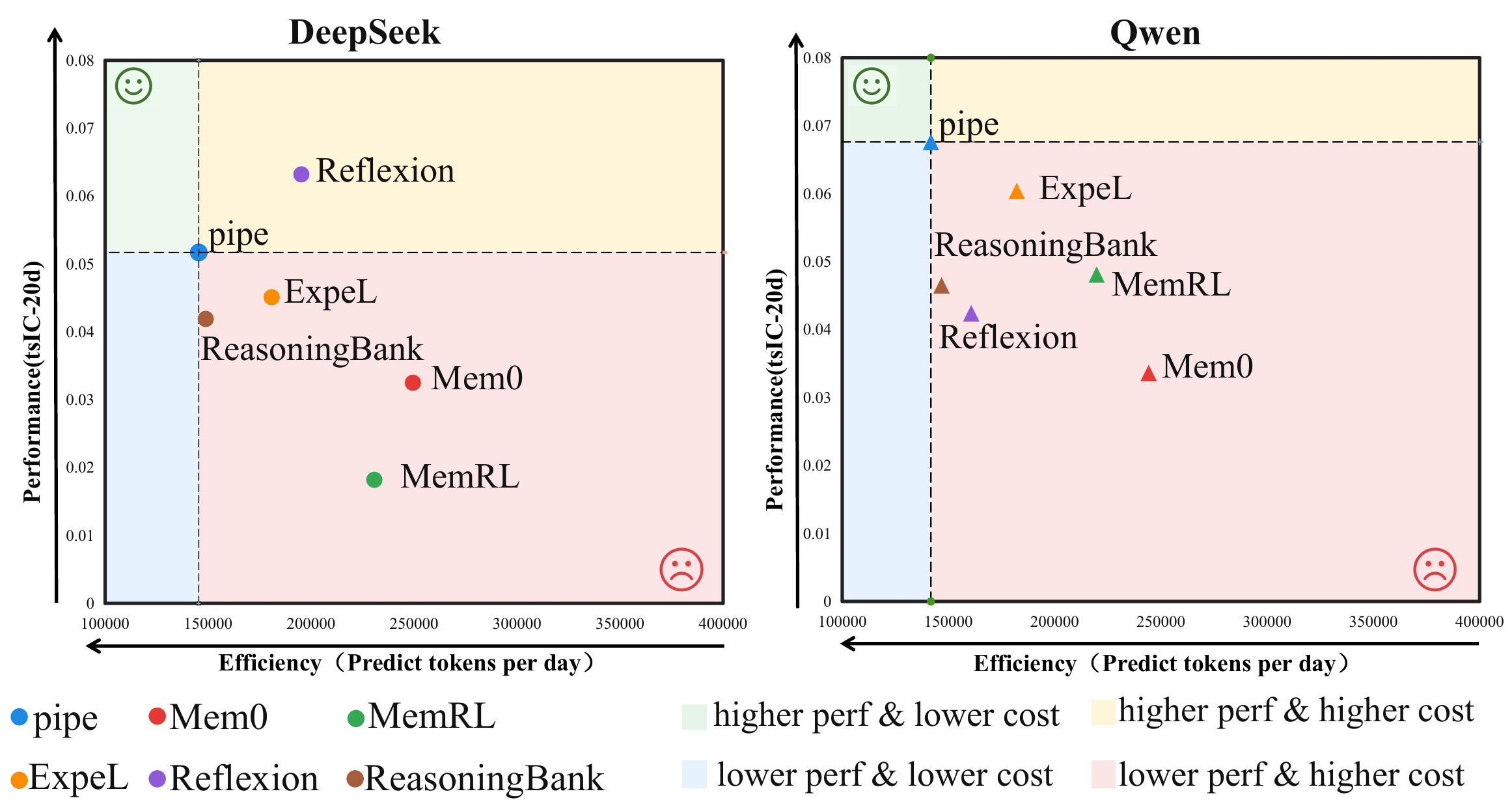}
    \caption{Performance--efficiency landscape of experience mechanisms on \textsc{FinEvolveBench}.
Points denote different experience methods under two LLM backbones.
}
    \label{fig:fig0}
    \vspace{-1mm}
\end{figure}

Our review further shows that existing benchmarks often simplify at least one of these dimensions, leaving more challenging combinations of scenario properties underexplored. In real-world settings, however, several challenges may arise simultaneously. Financial prediction provides a natural example. Market outcomes are only partially informative about the quality of a prior decision because they are affected by latent macroeconomic, behavioral, and market factors. Credit assignment is difficult because multiple news events, contextual signals, and previously accumulated experiences may jointly influence a prediction. The environment is also highly non-stationary: the relationship between news and market response changes across time and market regimes. Finally, the open-ended nature of real-world financial news leads to highly diverse and weakly recurring events, making it difficult to retrieve closely matched historical cases and directly reuse past experience. Together, these properties create a challenging experience regime in which past experiences may be noisy, difficult to attribute, rapidly outdated, and only weakly reusable.

To address this gap, we introduce \textsc{FinEvolveBench}, a benchmark for evaluating experience-based self-evolution under challenging experience regimes. Unlike existing financial benchmarks that provide either isolated QA instances or market time series without rich raw news context, \textsc{FinEvolveBench} reconstructs a fixed chronological stream of 126,878 original news articles and subsequent market outcomes across 31 Chinese A-share industry indices. On each prediction date, an agent observes only currently available information, produces horizon-specific market-response predictions, and receives the corresponding outcome only after the prediction horizon matures. This design preserves the full temporal process of experience evolution, allowing experience to be accumulated, validated, revised, and reused over time under identical causal information constraints and without future leakage.

Our experiments further highlight the need for \textsc{FinEvolveBench}. As shown in Fig.~\ref{fig:fig0}, across representative experience and memory mechanisms, multiple prediction horizons, and different LLM backbones, experience augmentation does not consistently improve performance over the no-experience baseline. These results indicate that the effectiveness of current experience mechanisms are not yet well characterized under challenging real-world experience regimes, motivating further study of how experience should be acquired, updated, and reused under such conditions.

%%%%%%
Our contributions are threefold:
\begin{itemize}

\item \textbf{We provide a scenario-oriented analysis of experience mechanisms.}
We characterize representative evaluation settings along four dimensions---outcome observability, credit assignment complexity, environmental dynamics, and experience reusability---and analyze how these properties shape the challenges faced by experience mechanisms.

\item \textbf{We introduce \textsc{FinEvolveBench}, a benchmark for self-evolution under a challenging experience regime.}
The benchmark combines noisy outcome feedback, complex credit assignment, environmental non-stationarity, and limited experience reusability. Its causal replay protocol enables controlled evaluation of how external experience is accumulated, updated, and reused over time.

\item \textbf{We reveal the deficiency of existing experience mechanisms under a challenging experience regime.}
% Across representative methods, backbones, and prediction horizons, we find that experience-based gains are inconsistent and can even reverse under different operating conditions, highlighting experience validity under changing environments as a key unresolved challenge for self-evolving agents.
Within \textsc{FinEvolveBench}, experience-based gains vary substantially across methods, backbones, and prediction horizons, and can even reverse, highlighting experience validity under changing environments as a key challenge for self-evolving agents.
\end{itemize}

%本文件是从 Scenarios Developer 来的洁净版
\section{Scenario Taxonomy and Related Work}
\label{sec:related_work}

In this section, we provide a scenario-oriented taxonomy for experience-based LLM agents and review representative methods through this lens. Rather than organizing prior work solely by model or memory architecture, we characterize their evaluation scenarios along four dimensions---Outcome \underline{O}bservability (O), \underline{C}redit Assignment Complexity (C), Environmental \underline{D}ynamics(D) and Experience \underline{R}eusability (R)---to reveal their scenario-specific challenges and the corresponding design requirements for experience mechanisms.

\begin{table*}[h]
\centering
\vspace{-10pt}
\caption{Scenario Characteristics and Representative Experience Mechanisms for LLM Agents}
\label{tab:experience_scenarios}

% \resizebox{\textwidth}{!}{
% \begin{tabular}{
% p{2.0cm}
% p{0.8cm}
% p{0.8cm}
% p{0.8cm}
% p{0.8cm}
% p{3.2cm}
% p{5.8cm}
% }
\resizebox{\textwidth}{!}{
\begin{tabular}{
p{2.6cm}
p{1.0cm}
p{1.0cm}
p{1.0cm}
p{1.0cm}
p{4.0cm}
p{5.6cm}
}

\toprule
\textbf{Scenario} &
$\boldsymbol{O}$ &
$\boldsymbol{C}$ &
$\boldsymbol{D}$ &
$\boldsymbol{R}$ &
\textbf{Representative Datasets} &
\textbf{Representative Experience Methods} \\
\midrule

QA / Reasoning &
High &
Medium &
Low &
High &
HotpotQA, FEVER, MMLU &
Reflexion~\cite{shinn2023reflexion},MetaReflection~\cite{gupta2024metareflection},ExpeL~\cite{zhao2024expel},Dynamic Cheatsheet~\cite{suzgun2026dynamic}.
\\

\midrule

Web / GUI Interaction &
High &
High &
Low &
High &
WebArena, Mind2Web, WebShop, MiniWoB++ &
Autoguide~\cite{fu2024autoguide},Expel~\cite{zhao2024expel},Contextual Experience Replay (CER)~\cite{liu2025contextual},Agent Workflow Memory (AWM)~\cite{wang2024agent},ReasoningBank~\cite{ouyang2026reasoningbank}.
\\

\midrule

Coding / Software Engineering &
High &
Medium &
Low &
High &
SWE-bench, SWE-bench Lite &
ExpeRepair~\cite{mu2026experepair},ReasoningBank~\cite{ouyang2026reasoningbank},RepoMem~\cite{wang2026improving}.
\\

\midrule

Goal-Oriented Embodied Tasks &
High &
High &
Low &
High &
ALFWorld, ScienceWorld &
ICAL~\cite{sarch2024vlm},Reflexion~\cite{shinn2023reflexion},ExpeL~\cite{zhao2024expel},Self-Generated In-Context Examples~\cite{sarukkai2026self}
\\

\midrule

Open-World Continual Exploration &
Medium &
High &
Low &
Medium &
Minecraft, MineDojo &
Voyager~\cite{wang2023voyager}
\\

\midrule

Dialogue / Social Interaction &
Low &
High &
High &
Medium &
SOTOPIA, Lifelong-SOTOPIA, $\tau$-bench, LoCoMo &
Generative Agents~\cite{park2023generative},MemoryBank~\cite{zhong2024memorybank},SAMULE~\cite{ge2025samule}
\\

\midrule

Financial Sentiment Analysis &
Low &
High &
High &
Low &
Financial PhraseBank, SEntFiN, FinQA, TAT-QA, FinBen &
No published papers
\\

\bottomrule
\end{tabular}
}

\vspace{-1mm}
\footnotesize

\end{table*}

\textbf{Outcome Observability (O).}
Outcome Observability measures the extent to which the outcomes of an agent's actions are directly and reliably observable from the environment. High observability provides informative feedback on the consequences of an agent's decisions, enabling agents to extract higher-quality experience. Coding tasks are a representative example, where outcomes can often be objectively evaluated through execution or tests, and failures can be localized with relatively high precision using compiler messages, stack traces, or failed test cases.
Partial observability, in contrast, arises when action outcomes can only be imperfectly inferred. This typically stems from two sources. First, \emph{intrinsic outcome uncertainty} occurs when observed outcomes are jointly determined by the agent's actions and latent or stochastic environmental factors. For example, in financial decision-making, subsequent market movements provide only noisy evidence of whether a decision or strategy was effective, since returns are influenced by numerous unobserved factors. Second, \emph{evaluation uncertainty} arises when task outcomes lack objective success criteria and must be approximated by human or model-based evaluators. Open-ended social interaction is a representative case: SOTOPIA~\cite{zhou2024sotopia} employs LLM-based evaluators to assess dimensions such as goal achievement and relationship maintenance, while SOTOPIA-$\pi$~\cite{wang2024sotopia} further reveals potential biases in such model-based evaluations.
Under partial observability, individual outcomes may provide unreliable supervision, causing agents to learn spurious correlations or incorrectly attribute success and failure. Experience mechanisms therefore need to aggregate evidence across multiple episodes, estimate the reliability of extracted experience, and revise or discard experience as additional evidence accumulates. ExpeL~\cite{zhao2024expel}, for example, reduces single-trajectory bias by extracting and iteratively refining insights from collections of successful and failed trajectories, while Closing the Feedback Loop~\cite{cui2026closing} maintains persistent evidence for experiential rules and updates their reliability across episodes under noisy and non-stationary financial feedback.

\textbf{Credit Assignment Complexity (C).}
Credit Assignment Complexity measures the difficulty of attributing an observed outcome to the specific actions or experiences that causally contributed to it. This difficulty can arise from several sources. First, the underlying reasoning process may itself be complex, with the final outcome jointly determined by multiple interdependent reasons, making it difficult to isolate the contribution of any single step. Second, credit assignment generally becomes harder as the horizon and dependency depth of the action sequence increase, since more intermediate actions may plausibly affect the final outcome. Third, it also depends on the agent's capacity to intervene in the environment. In highly interactive settings, such as simulators or games, agents can perform controlled perturbations, retry alternative actions, or repeatedly interact with the environment to test competing attribution hypotheses. When intervention capacity is limited, however, agents must infer causal contributions primarily from passive trajectories, where correlated decisions and environmental variation make consequential actions difficult to distinguish from incidental ones.

Reliable credit assignment is critical to experience quality: incorrect attribution may lead to noisy or even misleading experience. Related work has explored this challenge from different perspectives. Reflexion~\cite{shinn2023reflexion} uses task feedback and failed trajectories to generate verbal reflections on what should be corrected in subsequent attempts. ExpeL~\cite{zhao2024expel} further contrasts successful and failed trajectories to identify consequential differences and extract reusable insights. More recent work considers finer-grained failure attribution. Who\&When~\cite{zhang2025agent} attempts to identify which agent and which decision step caused a failure, but its low step-level accuracy illustrates the difficulty of localizing errors from trajectories alone. DoVer~\cite{ma2026dover} goes one step further by modifying suspected failure points and re-executing the task to test whether the attribution is correct, providing a more direct way to validate failure hypotheses.

\textbf{Environmental Dynamics (D).}
Environmental Dynamics measures the stability of an environment's transition regularities: whether the relationship between an action and its resulting state remains invariant across time and interactions. 
In low-dynamic environments, such as static reasoning tasks or reproducible simulators, past experience can remain valid across trials, while in dynamic or non-stationary environments, previously useful experiences may become harmful. The central challenge, therefore, shifts from experience acquisition to experience maintenance: agents must determine when experience should be revised, reinforced, or forgotten. MemoryBank~\cite{zhong2024memorybank} supports continual memory updating and selective forgetting over long-term interactions, while ReMe~\cite{cao2026remember} introduces utility-based refinement to retain useful procedural experience and prune outdated or low-utility memories. Similarly, Mem0~\cite{chhikara2025mem0} dynamically updates and consolidates stored memories as new information arrives, allowing outdated or conflicting information to be revised over time.

\textbf{Experience Reusability (R).}
Experience Reusability measures the expected degree to which future tasks share transferable structures, procedures, strategies, or skills with previously encountered tasks. In high-reusability settings, prior experience can be directly reused or adapted with only minor modifications. Web/GUI interaction is a representative example where many tasks share recurring operational patterns such as navigation, search, filtering, form filling, and submission.

By contrast, in highly instance-specific or weakly recurrent settings, such as financial sentiment analysis where relevant signals and market conditions vary substantially across events, raw historical trajectories offer limited direct utility. The central challenge is therefore to abstract task-specific experience into transferable knowledge while preserving task-relevant distinctions and avoiding over-generalization or negative transfer. Existing methods enhance transferability by increasing the abstraction and compositionality of experience. MemRL~\cite{memrl2026} combines similarity retrieval with feedback-updated utility, prioritizing experiences that prove useful across recurrent tasks while down-weighting ineffective ones. ExpeL~\cite{zhao2024expel} aggregates successful and failed trajectories into task-independent natural-language insights, allowing learned strategies to generalize beyond individual episodes. Voyager~\cite{wang2023voyager} further encodes successful behaviors as executable and compositional skills that can be retrieved and recombined for novel tasks in open-world environments.

%We 将现有的常见XXXX总结到表XXX中，分High, medium, low给出了其在各个维度上的评分，并列出了现有的代表性数据集，以及在此数据集上进行evaluation的相关工作。
\textbf{Scenario Classification and Characterization.}
Table~\ref{tab:experience_scenarios} summarizes representative evaluation scenarios under this taxonomy. For each scenario, we qualitatively characterize the four dimensions as \emph{High}, \emph{Medium}, or \emph{Low}, and list representative datasets and experience mechanisms evaluated in these settings. 
As shown in Table~\ref{tab:experience_scenarios}, existing benchmarks evaluate experience and memory mechanisms across a broad range of settings, but they typically make at least one component of experience-based self-evolution comparatively easy. In QA/reasoning and coding scenarios, such as HotpotQA~\cite{yang2018hotpotqa}, FEVER~\cite{thorne2018fever}, and SWE-bench~\cite{jimenez2024swebench}, feedback is relatively clear and the underlying task distribution is stable, allowing past experience to remain valid.  Web/GUI and embodied scenarios, such as WebArena~\cite{zhou2024webarena}, Mind2Web~\cite{deng2023mind2web}, and ALFWorld~\cite{shridhar2020alfworld}, introduce longer-horizon decision making and more difficult credit assignment, yet successful workflows or strategies often remain reusable across similar tasks. Continual environments such as Minecraft, MineDojo~\mbox{\cite{fan2022minedojo}}, and Lifelong-SOTOPIA~\mbox{\cite{goel2025lifelong}} further introduce changing contexts and weaker feedback, but previously acquired skills or interaction patterns can still transfer across future situations. Consequently, benchmarks that jointly exhibit noisy feedback, difficult credit assignment, strong non-stationarity, and low experience reusability remain underexplored. \textsc{FinEvolveBench} targets this gap, where the key challenge is not merely to retain experience, but to determine what can be reliably learned, when it should be revised, and whether it remains applicable.

\paragraph{Financial Sentiment Analysis as a Challenging Scenario.}
Financial prediction naturally combines the four challenges identified above: market feedback is noisy, causal attribution is ambiguous, action--outcome relationships evolve over time, and closely matched historical cases are scarce. It therefore provides a useful setting for evaluating experience mechanisms when these difficulties arise jointly. Existing financial benchmarks, however, cannot readily support the study of this form of experience-based self-evolution. FinQA~\cite{chen2021finqa}, TAT-QA~\cite{zhu2021tat}, FinBen~\cite{finben2024}, and financial sentiment datasets primarily provide static, instance-level evaluation: their samples are evaluated independently and do not provide the temporally linked observations and subsequent market feedback needed to accumulate, update, and reuse experience across decisions. StockBench~\cite{chen2025stockbench} introduces temporal trading simulation, but its primary objective is to evaluate trading decisions and profitability rather than the behavior of experience mechanisms themselves. Moreover, its news context is retrieved dynamically through a search API for each stock and trading day, making it difficult to guarantee that different methods are evaluated under an identical and reproducible information stream. Existing financial benchmarks therefore either omit the process of experience evolution or lack the experimental control needed to isolate it.

To address these limitations, we introduce \textsc{FinEvolveBench}, a temporally grounded, news-rich and reproducible benchmark that enables systematic evaluation of experience-based self-evolution under noisy feedback, ambiguous credit assignment, non-stationary dynamics, and limited experience reusability.

\begin{figure}[h]
    \vspace{-20pt}
    \setlength{\abovecaptionskip}{3pt}
    \centering
   \raisebox{6pt}[\height][\depth]{%
    \includegraphics[width=\linewidth]{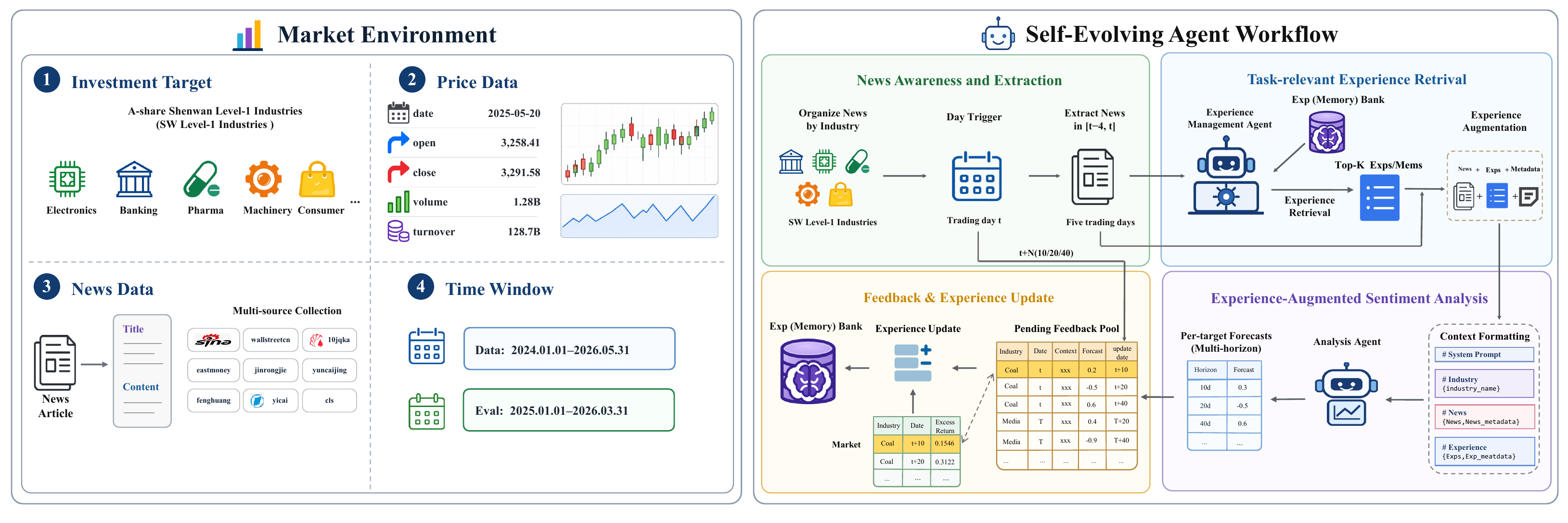}%
}
    \caption{\textbf{Overview of \textsc{FinEvolveBench}.}
    The benchmark reconstructs a chronological market environment from industry-level prices and multi-source financial news.
    At each trading day $T$, the agent retrieves task-relevant experience, predicts 10-, 20-, and 40-day market responses, and updates its experience only when the corresponding delayed outcomes mature, forming a causal loop of retrieval, prediction, feedback, and evolution without future leakage.}
    \label{fig:workflow_fig}
    \vspace{-10pt}
\end{figure}

\section{FinEvolveBench}
\label{sec:benchmark}

%先要说它是个什么数据集
% \textsc{FinEvolveBench} build on 一个金融市场行业情绪分析的数据集。On each trading day, an agent observes currently available financial news and market context, predicts how different industries will respond 特定时间之后. 在这个数据集之上，我们构建了XXXX（pipeline），where 
% LLM-agents can improve their performance by accumulating and updating external experience . 
% FinEvolveBench provides a fixed and reproducible stream of temporally aligned financial news and market outcomes, enabling controlled evaluation of experience accumulation, updating, and reuse over time.
\begin{wraptable}{r}{0.5\textwidth}
    \vspace{-22.5pt}
    \centering
    \setlength{\belowcaptionskip}{2pt}
    \caption{Core statistics of \textsc{FinEvolveBench}. The listed horizons describe our experiments rather than a restriction of the released data.}
    \label{tab:benchmark_stats}
    % \small
    \footnotesize
    \setlength{\tabcolsep}{4pt}
    \renewcommand{\arraystretch}{0.75}
    \resizebox*{!}{0.78\totalheight}{%
    \begin{tabular}{ll}
        \toprule
        \textbf{Statistic} & \textbf{Value} \\
        \midrule
        Market universe & 31 first-level industry \\
        News span & 2024-01-01 to 2026-06-01 \\
        Market-label span & through 2026-06-30 \\
        Prediction window & 2025-01-02 to 2026-03-31 \\
        Evaluation days & 300 trading days \\
        News articles & 126,878 \\
        Active news days & 876 \\
        Average news per day & 143.85 \\
        Experimental horizons & 10, 20, 40 trading days \\
        \bottomrule
    \end{tabular}%
}
\end{wraptable}
\textsc{FinEvolveBench} is a temporally ordered benchmark for evaluating
experience-based self-evolution in LLM agents.
It is built on an industry-level financial prediction task in which an agent
observes currently available financial news and market information, predicts
the future market response of different industries, and receives the
corresponding market outcome only after a predefined prediction horizon.
On top of this task, we construct a chronological replay pipeline that allows
agents to accumulate, update, and reuse external experience as feedback
becomes causally available.
Unlike static financial benchmarks or settings with dynamically retrieved
context, \textsc{FinEvolveBench} provides a fixed and reproducible stream of
temporally aligned news and market data, enabling controlled evaluation
of experience evolution over time.

% \begin{table}[h]
%     \caption{Core statistics of \textsc{FinEvolveBench}. The listed horizons describe our experiments rather than a restriction of the released data.}
%     \vspace{4pt}
%     \label{tab:benchmark_stats}
%     \centering
%     \small
%     \setlength{\tabcolsep}{4pt}
%     \begin{tabular}{@{}lp{0.64\textwidth}@{}}
%         \toprule
%         \textbf{Statistic} & \textbf{Value} \\
%         \midrule
%         Market universe & 31 first-level industry indices \\
%         News span & 2024-01-01 to 2026-06-01 \\
%         Market-label span & through 2026-06-30 \\
%         Prediction window & 2025-01-02 to 2026-03-31 \\
%         Evaluation days & 299 trading days \\
%         News articles & 126,878 \\
%         Active news days & 876 \\
%         Average news per day & 143.85 \\
%         Experimental horizons & 10, 20, 40 trading days \\
%         \bottomrule
%     \end{tabular}
% \end{table}

\subsection{Data Collection and Construction}
% 说一下我们数据的特点，“全面” 因为只有足够全面，feedback signal才能去做经验反思，说明一下意义，然后再具体说明Data Collection从Market universe 到XXX如何全面。
Experience-based self-evolution requires not only temporally ordered inputs, but
also sufficiently broad and consistently aligned observations from which
feedback can be interpreted.
We therefore construct a fixed financial information stream that combines
industry-level prediction targets, multi-source financial news, and subsequent
market outcomes over a continuous time period.

\textbf{News Corpus}
We collect 222,025 publicly accessible financial news articles from nine major Chinese media sources: Yuncaijing, ifeng Finance, Sina Finance, Eastmoney, Tonghuashun, JRJ, Wallstreetcn, Yicai, and Cailian Press. Each article is stored with its complete title, content, first-publication time, and source. We then use an LLM-based annotation pipeline to assign each article to the 31 Shenwan Hongyuan first-level industries in the Chinese A-share market and score its importance on a 1--10 scale according to predefined criteria. To reduce irrelevant or weakly informative inputs, we remove market-wide news and articles that cannot be assigned to any specific industry, and retain 126,878 articles with only industry-specific tags and an importance score of at least 8.

% We use the 31 Shenwan Hongyuan first-level industries in the Chinese A-share market to define the prediction universe and 

% Experience-based adaptation requires not only temporally ordered inputs, but
% also sufficiently broad and consistently aligned observations from which
% feedback can be interpreted. We collect 177,324 publicly accessible financial news articles from nine major Chinese media sources: Yuncaijing, ifeng Finance, Sina Finance, Eastmoney, Tonghuashun, JRJ, Wallstreetcn, Yicai, and Cailian Press.Each article is stored with its complete title, content, first-publication time, and source, and is further annotated with related-industry and importance metadata using an LLM-based pipeline. We select 31 Shenwan Hongyuan first-level industry indices from the Chinese A-share market 作为划分行业的依据，并且过滤了不重要的新闻
% 以减小对预测的影响XXXX。

% market data 说明
\textbf{Market Data}
% We collect XXXX market data 有哪些，怎么collect的，参考下一段。
We collect daily closing price series of 31 first-level industries under the Shenwan Hongyuan Industry Classification for the Chinese A Stock market.
Daily market records contain open and close prices, trading volume, and transaction fields; missing values remain \texttt{NaN} rather than being imputed. For a prediction made on date $t$, we compute the close-to-close return of industry $i$ over horizon $h$,
\begin{equation}
r^{i}_{t,h}=\frac{\mathrm{close}^{i}_{t+h}}{\mathrm{close}^{i}_{t}}-1,
\end{equation}
and subtract the same-horizon CSI1000 return:
\begin{equation}
\alpha_{i,t,h}=r^{i}_{t,h}-r^{\mathrm{CSI1000}}_{t,h}.
\label{eq:excess_return}
\end{equation}
This market adjustment separates an industry's relative response from broad market movement. The resulting $\alpha_{i,t,h}$ is used as delayed feedback and as the evaluator-side outcome for the prediction score $s_{i,t,h}$.

% 这里可以提一下，就是相关数据总结在表二上，引用一下 Table 2，不然这个 Table 好像没有被引用
The experimental replay contains 300 prediction days from January 1, 2025 through March 31, 2026, with the complete dataset statistics summarized in Table~\ref{tab:benchmark_stats}. The released records support other researcher-defined horizons within their temporal coverage. Market observations extend through June 30, 2026 so that pending outcomes near the end of our experimental window can be computed without exposing those future observations to the agent.

\begin{figure}[h]
    \vspace{-10pt}
    \centering
    \includegraphics[width=0.9\textwidth]{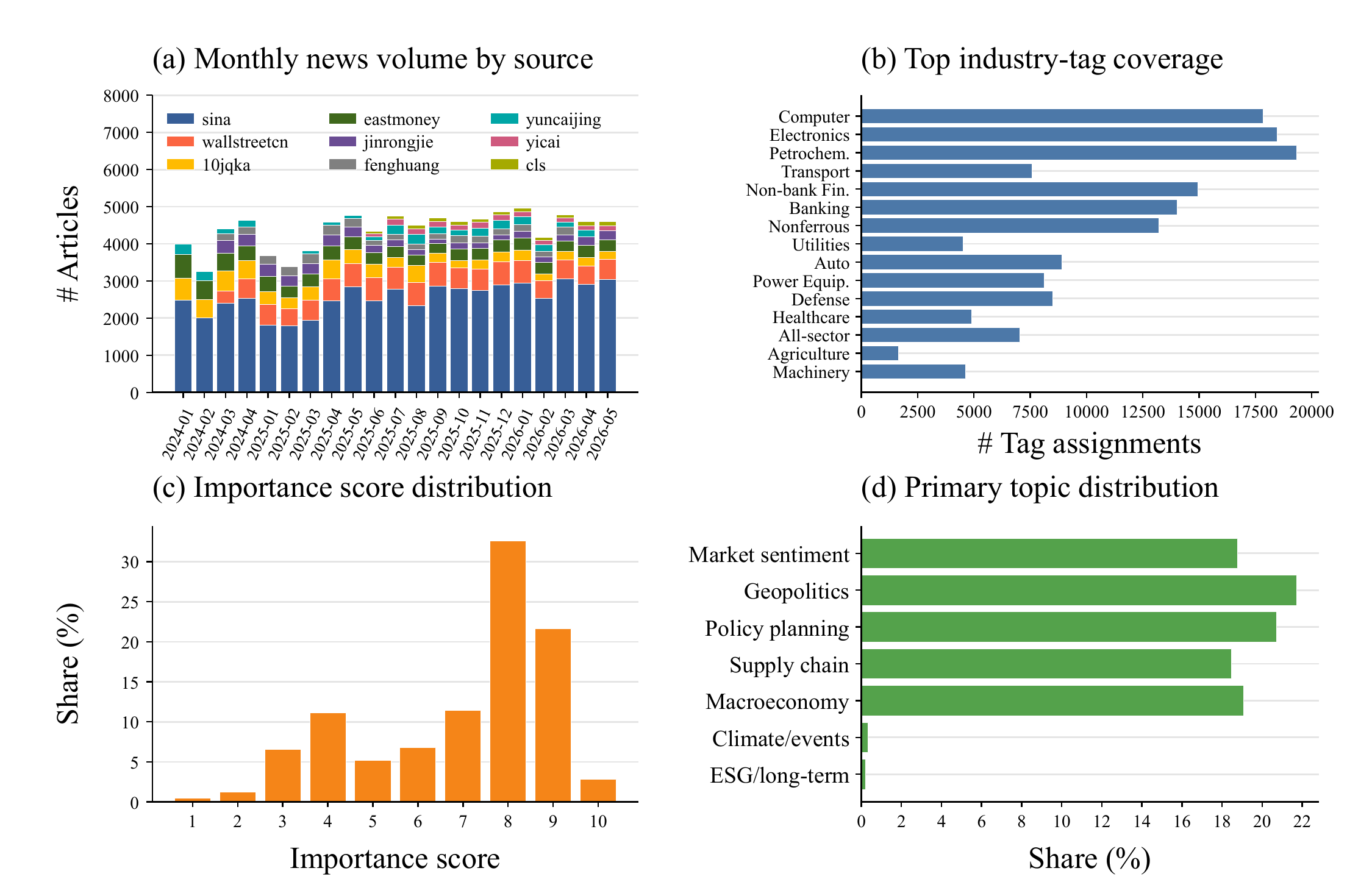}
    \caption{News corpus profile, including monthly source volume, industry-tag coverage, importance scores, and topic distributions.}
    \label{fig:news_profile}
    \vspace{-10pt}
\end{figure}

\subsection{Problem Formulation and Evaluation Protocol}

% 我修改了顺序，所以一些东西如价格变化率已经定义好了，可以做适当修改
As shown in Fig.~\ref{fig:workflow_fig}, for each industry $i$ on prediction date $t$, the agent observes financial news published from $t-4$ to $t$, the latest market data, and all experience available before $t$. Articles published on weekends or holidays are included in the next applicable prediction window according to their original publication dates. Based on this information, the agent produces a continuous market-response score $s_{i,t,h}\in[-1,1]$ for each horizon $h\in\mathcal{H}$, representing the expected direction and magnitude of the industry's future price change.

At time $t+h$, the realized industry return becomes observable and serves as delayed feedback for the prediction made at $t$. The agent then uses this feedback to update or generate experience, which becomes available for subsequent tasks.

\subsection{Dataset Statistics and Analysis}
%这边看着表和表例，靠得有点太近

% As shown in Figure~\ref{fig:news_profile}, our collected news corpus provides dense, continuous coverage across the replay period. As shown in Figure~\ref{fig:news_profile}, monthly news volume remains substantial across sources, while industry exposure is long-tailed.
As shown in Figure~\ref{fig:news_profile}, the collected news corpus provides a sustained multi-source financial news stream, with thousands of articles collected monthly from nine media sources. 
The collected news spans diverse industries and market themes, providing heterogeneous contexts for chronological experience evaluation.
Sina Finance contributes 55.99\% of articles, and 56.63\% of articles receive an importance score of at least 8. The topic distribution covers market sentiment, geopolitics, policy planning, supply chains, macroeconomics, climate and events, and longer-term themes rather than a single financial-news category.

\section{Experiments}
\label{sec:experiments}

In this section, we evaluate representative experience mechanisms on \textsc{FinEvolveBench} to investigate their effectiveness under the joint challenges identified in Section~\ref{sec:related_work}. Specifically, we study whether experience mechanisms consistently improve over an experience-free pipeline across different operating conditions, including prediction horizons and LLM backbones.

\subsection{Experimental Setup}

We evaluate our method separately with \textsc{DeepSeek-V4-Flash} and \textsc{Qwen3.6-35B-A3B} to examine whether the observed results generalize across model backbones.
Both models use a 32,768-token context window and their recommended decoding configurations.
Experiments are conducted at prediction horizons of 10, 20, and 40 trading days.
We report both time-series information coefficient (tsIC) and cross-sectional information coefficient (csIC), which respectively measure temporal predictive association within industries and same-day ranking quality across industries. Formal definitions are provided in Appendix~\ref{sec:evaluation_definitions}.

% \paragraph{Compared systems.}
We select five representative experience mechanisms that collectively span the major paradigms for constructing, maintaining, updating, reflecting on, and abstracting experience, and evaluate how they perform under the challenging regime instantiated by \textsc{FinEvolveBench}.
Pipe provides the experience-free reference, using only the currently available news and market context. We evaluate five representative experience mechanisms: Mem0 for dynamic agentic memory management, MemRL for feedback-calibrated memory with outcome-based utility updates, Reflexion for verbal reflection from interaction feedback, and ExpeL/ReasoningBank for abstraction-based experience that distills reusable insights or reasoning strategies from past trajectories. 
%每种方法都有针对我们在 Section 2 中提出的某种维度挑战，我们把它列在表二的 Challenge 那一栏，可以看到，我们选择的方法覆盖了我们提出的那四种挑战。
Each method is associated with one or more of the scenario challenges introduced in Section~\ref{sec:related_work}, which we summarize in the \emph{Challenges} column of Table~\ref{tab:overall_experience_results}. Collectively, the selected methods cover all four dimensions in our taxonomy, providing a diverse testbed for examining how different experience mechanisms behave under these challenges.
%Together, they span trajectory reuse, feedback updating, reflection, and experience abstraction under identical chronological and causal constraints.

\subsection{Performance of Existing Experience Mechanisms on FinEvolveBench}

\begin{table*}[t]
\vspace{-10pt}
\centering
\caption{
Overall performance of existing experience methods on \textsc{FinEvolveBench}.
$\uparrow$ indicates that an experience-augmented method outperforms the corresponding
Pipe baseline under the same backbone, horizon, and metric.
$\Delta$ denotes the average improvement over \textsc{Pipe} across the three horizons.
Higher is better for all metrics. The Challenges column denotes the scenario dimensions primarily addressed by each experience mechanism:
O,C,D,R.
}
\label{tab:overall_experience_results}

\small
\setlength{\tabcolsep}{4.2pt}
\renewcommand{\arraystretch}{1.08}

\resizebox{\textwidth}{!}{
\begin{tabular}{llcccccccc}
\toprule
& & \multicolumn{2}{c}{\textbf{10d}}
& \multicolumn{2}{c}{\textbf{20d}}
& \multicolumn{2}{c}{\textbf{40d}}
& \multicolumn{2}{c}{\textbf{Avg. Gain}} \\
\cmidrule(lr){3-4}
\cmidrule(lr){5-6}
\cmidrule(lr){7-8}
\cmidrule(lr){9-10}

\textbf{Method}
& \textbf{Challenges}
& \textbf{tsIC} & \textbf{csIC}
& \textbf{tsIC} & \textbf{csIC}
& \textbf{tsIC} & \textbf{csIC}
& $\Delta$\textbf{tsIC} & $\Delta$\textbf{csIC} \\
\midrule

\multicolumn{10}{l}{\textit{Panel A: DeepSeek-V4-Flash}} \\
% \addlinespace[2pt]

Pipe
& -- 
& 0.0243 & 0.0177
& 0.0517 & 0.0408
& 0.0595 & 0.0323
& -- & -- \\

\quad + Mem0
& D
& 0.0202 & 0.0149
& 0.0325 & 0.0240
& 0.0288 & $0.0416^{\uparrow}$
& -0.0180 & -0.0034 \\

\quad + MemRL
& D+O
& 0.0175 & 0.0147
& 0.0182 & 0.0060
& 0.0269 & 0.0264
& -0.0243 & -0.0146 \\

\quad + Reflexion
& D+C+O
& 0.0195 & $0.0223^{\uparrow}$
& $0.0632^{\uparrow}$ & $0.0525^{\uparrow}$
& $0.0623^{\uparrow}$ & 0.0162
& $+0.0032$ & $+0.0001$ \\

\quad + ExpeL
& C+R
& $0.0295^{\uparrow}$ & $0.0330^{\uparrow}$
& 0.0451 & 0.0280
& 0.0522 & 0.0094
& -0.0029 & -0.0068 \\

\quad + ReasoningBank
& C+R
& 0.0124 & 0.0091
& 0.0419 & 0.0285
& $0.0699^{\uparrow}$ & $0.0420^{\uparrow}$
& -0.0038 & -0.0037 \\

\midrule

\multicolumn{9}{l}{\textit{Panel B: Qwen3.6-35B-A3B}} \\
% \addlinespace[2pt]

Pipe
& -- 
& 0.0349 & 0.0342
& 0.0676 & 0.0542
& 0.0677 & 0.0452
& -- & -- \\

\quad + Mem0
& D
& 0.0132 & 0.0098
& 0.0336 & 0.0272
& 0.0475 & 0.0250
& -0.0253 & -0.0239 \\

\quad + MemRL
& D+O
& 0.0186 & 0.0135
& 0.0481 & 0.0514
& 0.0135 & -0.0065
& -0.0300 & -0.0251 \\

\quad + Reflexion
& D+C+O
& 0.0283 & $0.0407^{\uparrow}$
& 0.0424 & 0.0418
& 0.0452 & 0.0072
& -0.0181 & -0.0146 \\

\quad + ExpeL
& C+R
& 0.0262 & $0.0360^{\uparrow}$
& 0.0604 & $0.0607^{\uparrow}$
& 0.0441 & 0.0044
& -0.0132 & -0.0108 \\

\quad + ReasoningBank
& C+R
& 0.0101 & 0.0074
& 0.0465 & 0.0322
& $0.1033^{\uparrow}$ & $0.0716^{\uparrow}$
& -0.0034 & -0.0075 \\

\bottomrule
\end{tabular}
}
\vspace{-15pt}
\end{table*}

Table~\ref{tab:overall_experience_results} reports the overall tsIC and csIC of existing experience methods under the unified 10/20/40-day horizon. On \textsc{Dpsk-V4-Flash}, improvements over Pipe are sparse: ExpeL improves both metrics at 10 days, Reflexion improves both metrics at 20 days and tsIC at 40 days, while Mem0 improves only 40-day csIC. None of the existing methods consistently outperforms Pipe across horizons and metrics.ReasoningBank shows limited short-horizon gains but stronger improvements at 40 days, indicating that reasoning reuse may benefit longer-term prediction.

The pattern is even clearer on \textsc{Qwen3.6-35B-A3B}. Mem0 and MemRL remain below Pipe in all reported settings; Reflexion improves only 10-day csIC, while ExpeL improves csIC at 10 and 20 days but not tsIC. Accordingly, all five methods have negative average gains on Qwen, while on Dpsk only Reflexion achieves a marginally positive average gain ($+0.0032$ tsIC and $+0.0001$ csIC).

These results contrast with the gains previously reported for these methods across diverse experience-learning benchmarks. Rather than suggesting that the methods are generally ineffective, the results show that the compound challenges instantiated by \textsc{FinEvolveBench} substantially reduce the gains achieved by existing experience mechanisms, establishing \textsc{FinEvolveBench} as a more demanding testbed for experience-based self-evolution.

%\subsection{Sensitivity to Operating Conditions}
Moreover, the improvements provided by existing experience mechanisms are not consistent across different prediction horizons.
%We next examine whether the gains of existing experience mechanisms persist across prediction horizons. 
As shown in Table~\ref{tab:overall_experience_results}, changing the horizon can substantially alter both the magnitude and direction of their improvements over Pipe. For example, on \textsc{Dpsk-V4-Flash}, Reflexion improves csIC by \(+0.0046\) and \(+0.0117\) at 10 and 20 days, respectively, but degrades it by \(0.0161\) at 40 days. A stronger reversal appears for ExpeL on \textsc{Qwen3.6-35B-A3B}: its csIC gain increases from \(+0.0018\) at 10 days to \(+0.0065\) at 20 days, but drops to \(-0.0408\) at 40 days.

\subsection{Case Study: When Experience Becomes Misleading}
\label{subsec:case_study}
In this subsection, we provide a case study illustrating a failure mode of analyzed experience methods on \textsc{FinEvolveBench}. We examine a representative example from the transportation sector on May 30, 2025. 

 The experience-free pipeline predicts negative market sentiment across all horizons, with scores of $-0.10$, $-0.20$, and $-0.30$ for the 10-, 20-, and 40-day forecasts, respectively. These predictions align with the subsequent market movements, where the realized excess returns are $-2.37\%$, $-6.58\%$, and $-10.75\%$. 
In contrast, ExpeL, Reflexion, Mem0, and MemRL all produce positive predictions, opposite to the observed market direction. 
%This example illustrates that incorporating historical experience does not necessarily improve decision quality under challenging experience regimes; instead, improperly retained or applied experience may bias predictions away from current market conditions.
Additional details, including the input news and market context, retrieved experiences, prediction outputs, and detailed analysis of each method, are provided in Appendix~\ref{sec:case_study_appendix}. 

\textbf{Failure Analysis.}
Based on the case study and the designs of the evaluated mechanisms, we analyze their failure modes under the \textsc{FinEvolveBench} setting from two stages of the experience lifecycle: \emph{experience formation} and \emph{experience reuse}.

\emph{Experience formation.}
In complex financial sentiment analysis, extracting transferable and reliable experience is itself challenging. Mem0 and MemRL mainly preserve and retrieve past trajectories or memories with limited abstraction of causal factors. As shown in Appendix~\ref{sec:case_study_appendix}, their stored experiences remain close to the original news and prediction trajectories. However, market outcomes cannot be directly attributed to individual news events due to multiple latent factors and noisy feedback. Consequently, retrieved experiences may contain limited information or even encode misleading correlations. Reflexion and ExpeL perform explicit experience abstraction and generally achieve stronger performance. Nevertheless, abstraction alone does not guarantee valid experience. In the studied case, both methods extract bullish patterns from previous episodes, such as the relationship between tariff relief and shipping recovery, and reinforce these patterns in future predictions. However, such relationships may not persist under changing market conditions, causing unstable correlations to be encoded as reusable experience.

\emph{Experience reuse.}
Even when useful experience has been extracted, effectively retrieving and applying it remains challenging. Semantic similarity does not necessarily imply decision relevance. For example, Mem0 retrieves a highly similar transportation episode (similarity score $0.9353$), but the retrieved experience provides limited guidance because the original outcome may have depended on unobserved factors rather than the retrieved news pattern. Moreover, retrieved experience may bias predictions toward previous reasoning patterns despite conflicting current evidence. In this case, the current instance contains substantial negative signals, including a $19.8\%$ decline in U.S. goods imports, weakening port activity, and the possibility that previous shipping demand was merely pulled forward. Although ExpeL and Reflexion identify these signals, retrieved bullish experiences still dominate their predictions.

Overall, this case demonstrates that effective self-evolution in \textsc{FinEvolveBench} requires mechanisms that can evaluate experience validity, distinguish causal evidence from spurious correlations, and revise or discard outdated experience under non-stationary environments.

\section{Conclusion}
\enlargethispage{2\baselineskip}

We introduced \textsc{FinEvolveBench}, a benchmark for evaluating
experience-based self-evolution under challenging combinations of four
scenario properties: low outcome observability, high credit assignment
complexity, dynamic environments, and limited experience reusability. Beyond providing a new benchmark, we presented a systematic scenario-oriented review of experience mechanisms and showed that the effectiveness of experience-based adaptation should be interpreted together with the properties of the environment in which experience is acquired and reused.
Our experiments show that existing experience mechanisms do not consistently
improve over the experience-free pipeline in this setting. Their performance
changes across methods, backbones, prediction horizons,
with some experience mechanisms even introducing negative transfer. 
We hope \textsc{FinEvolveBench} serves as a controlled testbed for studying experience validity under challenging experience regimes, and motivates future experience mechanisms that enable agents to reliably evaluate, revise, and reuse experience over time.

% We introduced \textsc{FinEvolveBench}, a benchmark for evaluating
% experience-based self-evolution under challenging combinations of four
% scenario properties: low outcome observability, high credit assignment
% complexity, dynamic environments, and limited experience reusability.
% We further provided a scenario-oriented analysis of experience mechanisms,
% showing that their effectiveness depends on the environment in which
% experience is acquired and reused.

% Experiments demonstrate that existing mechanisms do not consistently
% outperform experience-free pipelines: their benefits vary across methods,
% backbones, and prediction horizons, with some causing negative transfer.
% We hope \textsc{FinEvolveBench} provides a controlled testbed for studying
% experience validity and inspires future agents that can reliably evaluate,
% revise, and reuse experience over time.

\section*{AI Use Statement}
\label{sec:ai_use_statement}
%太长，空间宝贵，两三句话解决
In this work, we used LLMs to assist with specific software implementation details and translation. The overall software framework and methodological logic were developed by the authors, while LLMs were used to assist with the implementation of selected code components. All LLM-assisted code was manually reviewed by the authors to verify its logic and correctness. For translation, LLMs were used to translate author-written Chinese text into preliminary English drafts, which were subsequently reviewed and revised by the authors.

We did not use LLMs to generate synthetic datasets, develop theoretical models or conceptual frameworks, formulate mathematical claims, assist with proofs, propose or refine hypotheses, design research methodology or experiments, clean or reformat datasets, conduct qualitative or thematic data analysis, or interpret experimental results. LLMs were also not used to generate citations or references.

All LLM-assisted work was reviewed by the authors. We take responsibility for the final content of this work, including text, claims, code, and other artifacts produced with the aid of LLMs.

\section*{Ethics Statement}
\label{sec:ethics_statement}

\textsc{FinEvolveBench} evaluates representative experience mechanisms that
span major paradigms of experience construction and reuse, including
trajectory reuse, feedback-calibrated memory, reflection, and
cross-trajectory abstraction. However, the benchmark does not aim to
exhaustively characterize all possible designs of self-evolving agents.
Therefore, the observed behaviors should be interpreted as evidence about
representative mechanism families under the evaluated scenarios, rather than
as a complete characterization of future experience-based agent designs.

The benchmark is constructed from Chinese A-share industry indices and
Chinese financial news, and its findings may not directly generalize to other
markets, asset classes, or languages. In addition, the benchmark pipeline
relies on upstream news collection and LLM-based metadata annotation.
Although these annotations are used only for organizing benchmark data,
potential collection gaps or annotation errors may introduce uncertainty into
downstream evaluations.

\textsc{FinEvolveBench} is designed to study the adaptation capability of LLM
agents in a temporally evolving financial environment, rather than to provide
investment recommendations or deployable trading systems. The reported
metrics, including tsIC and csIC, measure predictive associations with future
market outcomes and should not be interpreted as realized trading performance,
as practical investment decisions additionally depend on factors such as
transaction costs, liquidity, risk management, and portfolio construction.
The benchmark uses publicly accessible financial news and market data without
involving private user information or personal data. We emphasize that
benchmark results should be interpreted within the defined evaluation setting
and should not be regarded as financial advice or guarantees of future market
behavior.

\section*{Reproducibility Statement}

We will release the complete benchmark package upon acceptance, including the replay environment, baseline implementations, and evaluation scripts. 

To support reproducible evaluation, FINEVOLVEBENCH provides a fixed
chronological stream of financial news and market observations, together with a
standardized replay protocol. All evaluated systems share identical information
access, prediction horizons, backbone-specific decoding configurations, input
formats, and evaluation functions. In particular, the benchmark strictly
enforces causal feedback availability: future news, future market observations,
and unresolved outcomes are inaccessible before prediction; completed
trajectories are appended only after prediction and become available for future
retrieval. These controls prevent information leakage and ensure that different
experience mechanisms are compared under the same temporal constraints.

The remaining sources of variation mainly arise from the underlying language
models. Our experiments involve LLM backbones whose outputs may vary due to
model updates, API availability, and inherent stochasticity in generation.
Although we fix prompts, decoding configurations, and evaluation procedures to
reduce such variation, exact token-level reproduction may become difficult if
the underlying models are modified or retired by providers. We therefore
consider the released benchmark environment and evaluation protocol, rather
than a single model execution trace, as the primary reproducibility target.

Finally, FINEVOLVEBENCH is constructed from Chinese A-share industry indices and
financial news collected from a specific market environment. While the released
benchmark enables consistent reproduction within this setting, conclusions
should be interpreted with respect to this environment and may not directly
transfer to other markets, asset classes, or languages.

\bibliography{main}
\bibliographystyle{iclr2027_conference}

\clearpage
\appendix
\section{Additional Benchmark Details}
\label{sec:appendix}

\subsection{Investment Universe}

\begin{figure*}[h]
    \centering
    \includegraphics[width=0.65\textwidth]{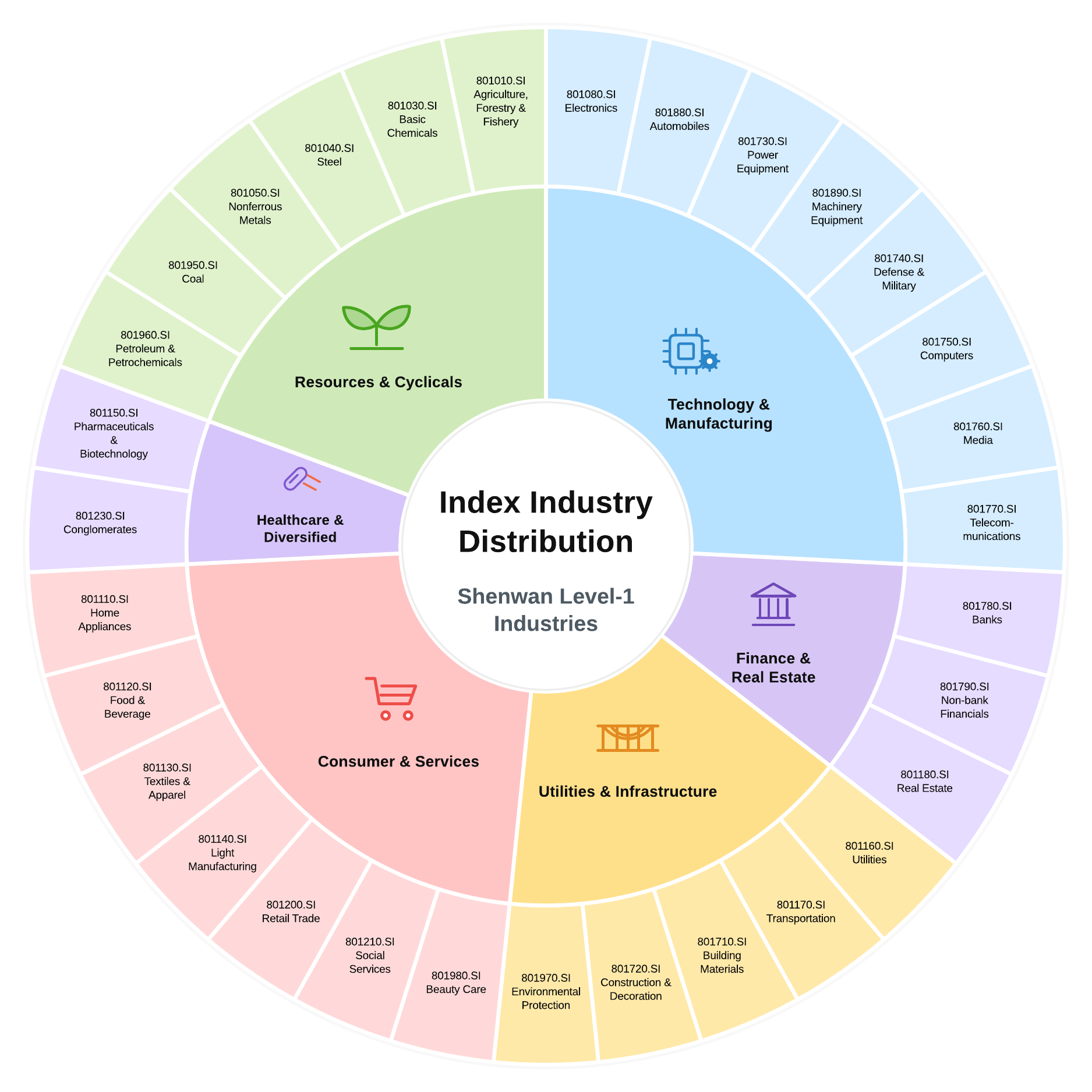}
    \caption{Industry coverage of the selected first-level indices.}
    \label{fig:industry}
\end{figure*}

Figure~\ref{fig:industry} shows the coverage of the 31 Shenwan Hongyuan first-level industry indices. The universe spans major sectors and supports evaluation across heterogeneous market narratives.

\subsection{Collection and Data Format}

For each source, we collect only publicly accessible content that requires neither payment nor registration. The crawler records title, body text, publication time, and source website while respecting each site's \texttt{robots.txt} rules and rate limits.

News and market observations are stored in daily \texttt{YYYYMMDD.csv} files. A task record contains \textit{date}, \textit{sector}, \textit{newsList}, \textit{open}, \textit{close}, \textit{volume}, and transaction-related fields. Missing price fields are retained as \texttt{NaN}.

\subsection{Detailed Chronological Replay Protocol}

Figure~\ref{fig:overview} summarizes the daily replay sequence used in our experiments.

The benchmark standardizes the following order without prescribing the internal design of an experience mechanism:
\begin{enumerate}
    \item \textbf{Mature and update:} after the close, reveal outcomes of earlier predictions whose horizons end on the current trading day. Eligible systems update only the corresponding horizon-specific utility before the current prediction.
    \item \textbf{Observe:} at the 24:00 cutoff, expose the current close, date-bounded news, and all earlier causally available information. News is indexed by its first publication time.
    \item \textbf{Retrieve and predict:} optionally retrieve memory that existed before the current prediction, then produce and freeze the score for each selected horizon.
    \item \textbf{Append and queue:} append the completed prediction trajectory to eligible memory stores and queue its outcomes. The new trajectory becomes retrievable only from the next trading day; its utility can change only when the corresponding delayed outcome matures.
\end{enumerate}

Systems without experience skip retrieval, trajectory storage, and utility updating. Append-only systems store and retrieve complete prediction trajectories but never attach later outcomes or update their utility. Feedback-updated systems follow the same append timing and may update only horizon-specific utility after the corresponding outcome matures. All systems retain identical task order and information access.

\begin{figure}[h]
    \centering
    \small
    \fbox{\parbox[c][2.7cm][c]{0.19\textwidth}{\centering
        \textbf{1. Mature}\\[2pt]
        Reveal eligible prior\\
        outcomes; update utility}}
    \hfill$\longrightarrow$\hfill
    \fbox{\parbox[c][2.7cm][c]{0.19\textwidth}{\centering
        \textbf{2. Observe}\\[2pt]
        Date-bounded news\\
        and market context}}
    \hfill$\longrightarrow$\hfill
    \fbox{\parbox[c][2.7cm][c]{0.19\textwidth}{\centering
        \textbf{3. Retrieve/Predict}\\[2pt]
        Use causally available\\
        memory; freeze scores}}
    \hfill$\longrightarrow$\hfill
    \fbox{\parbox[c][2.7cm][c]{0.19\textwidth}{\centering
        \textbf{4. Append/Queue}\\[2pt]
        Store trajectory; queue\\
        selected-horizon outcomes}}
    \caption{Daily chronological replay sequence. The same causal order is applied to every system; only the presence, representation, retrieval, and updating of external experience may differ.}
    \label{fig:overview}
\end{figure}

\subsection{Evaluation Definitions}
\label{sec:evaluation_definitions}

Let $\mathcal{D}=(d_1,\ldots,d_T)$ be the shared chronological sequence of 300 evaluation trading days. We define
\begin{align}
\mathcal{D}_{\mathrm{cold}}
    &=\{d_1,\ldots,d_{\lfloor T/3\rfloor}\},\\
\mathcal{D}_{\mathrm{exploit}}
    &=\{d_{\lfloor T/3\rfloor+1},\ldots,d_T\}.
\end{align}
The first period measures behavior while experience is limited or being calibrated, and the second measures behavior after a longer history of interactions and matured feedback is available. \emph{Overall} denotes the complete sequence. These stages do not assume monotonic improvement. Correlations are recomputed within each stage; Cold-Start scores are not obtained by subtracting Exploitation scores from Overall scores because IC is not additive across time intervals.

Cross-sectional IC measures same-day ranking across targets:
\begin{equation}
\mathrm{csIC}_{t,h}=\operatorname{Corr}_{i}(s_{i,t,h},\alpha_{i,t,h}),
\end{equation}
and $\mathrm{csIC}_{h}=T^{-1}\sum_t \mathrm{csIC}_{t,h}$. Time-series IC measures temporal prediction for each target:
\begin{equation}
\mathrm{tsIC}_{i,h}=\operatorname{Corr}_{t}(s_{i,t,h},\alpha_{i,t,h}),
\end{equation}
and $\mathrm{tsIC}_{h}=N^{-1}\sum_i \mathrm{tsIC}_{i,h}$. We report both metrics by backbone, horizon, and temporal stage.

\subsection{Reproducibility Controls}

All systems share the same backbone-specific decoding settings, structured input metadata, chronological replay order, experimental 10/20/40-day prediction horizons, and evaluation functions. Every experience store is empty before the first evaluation date. At date $t$, feedback that matures after the close may update the corresponding MemRL utility before prediction; no method can access later news, later prices, or feedback for a still-pending prediction. Prediction uses news first published by the 24:00 cutoff and memory already available before the current prediction. The completed trajectory is appended only after prediction and becomes retrievable on the next trading day. Future market observations used to score predictions near the end of the window remain evaluator-only labels after they mature; no evaluated post-window action or memory update is performed. Outputs that cannot be parsed into valid horizon-specific scores are excluded only from that horizon, and metrics use the valid evaluation intersection. Hyperparameters are fixed before the replay and are not tuned using future evaluation outcomes.

We recommend reporting the full $2\times3$ metric matrix---tsIC and csIC at 10, 20, and 40 trading days---for every evaluated backbone. Results should additionally identify whether they use the Cold-Start, Exploitation, or Overall period. Reporting only the best horizon or a single terminal mean can hide negative transfer and unstable rankings.

% \subsection{Append-Only and Utility-Updated Memory Controls}

% Pipe+mem0 starts empty, appends each complete prediction trajectory, and retrieves prior trajectories semantically; it never attaches later outcomes or updates utility. Pipe+MemRL follows the same causal append timing but updates horizon-specific utility only after delayed feedback matures. Pipe+Frozen MemRL retains the same empty start, trajectory-writing procedure and timing, representation, and retrieval procedure as the feedback-updated version while suppressing only feedback-driven utility updates. Consequently, Updated--Frozen MemRL is the controlled comparison for utility-update effects; Mem0--MemRL is only a comparison between two different experience mechanisms.

\section{Implementation Details}

\subsection{Unified Experience Framework}

To enable a controlled comparison among different experience mechanisms,
FinEvolveBench abstracts experience-based self-evolution into three
standardized interfaces: experience retrieval, experience augmentation, and
experience update. All evaluated methods share the same prediction pipeline,
backbone model, input format, prediction horizons, and chronological replay
protocol; they differ only in how experiences are represented, retrieved,
augmented, and updated.

At each prediction date, the agent first receives available financial
news and experiences accumulated before the
current prediction. The experience mechanism retrieves relevant historical
information and integrates it into the prediction context. After producing
predictions for the 10-, 20-, and 40-day horizons, the trajectories are stored
until the corresponding market outcomes become available. Once delayed feedback
matures, the experience update procedure is executed and the updated experience
becomes available for subsequent predictions.

This unified interface preserves the original design principles of different
experience mechanisms while adapting them to the temporally ordered and
feedback-delayed setting of FinEvolveBench.

\subsection{Hyperparameter Configuration}

All hyperparameters are fixed before evaluation and are not tuned using future
evaluation outcomes. Reflexion uses a maximum reflection length of 4,000
characters and truncates news inputs during reflection updates. ExpeL retrieves
the latest 10 insights for each prediction. Mem0 retrieves up to 5 memories
with a similarity threshold of 0.1 and uses Qwen3-Embedding-8B as the embedding
model. MemRL retrieves 20 candidate experiences, selects the final 5 based on
utility ranking, and updates Q-values with a learning rate of 0.3.

\subsection{Reflexion}

We implement Reflexion as a sector-specific verbal reflection mechanism.
Each first-level industry maintains an independent reflection memory. During
prediction, the system retrieves the most recent available reflection for the
target industry and appends it to the current financial context. If no prior
reflection exists, prediction is performed using only the current news.

After a prediction outcome becomes available, the system updates the reflection
memory by generating a new reflection based on the historical reflection,
prediction result, realized market feedback, and corresponding news context.
The generated reflection summarizes previous reasoning, potential error causes,
and considerations for future predictions.

Since the original Reflexion framework is primarily designed for interactive
tasks with immediate environmental feedback, we adapt its feedback signal to
the delayed market feedback setting of FinEvolveBench. Specifically, reflection
updates are triggered only when the corresponding prediction horizon matures,
while preserving the original verbal reflection paradigm.

\subsection{ExpeL}

We implement ExpeL by converting completed prediction trajectories into
reusable natural-language insights. A trajectory consists of the prediction
date, prediction horizon, prediction score, realized market signal, and the
corresponding financial news context. Once the delayed outcome becomes
available, the trajectory is summarized into reusable insights, which are
stored in the sector-specific experience repository.

During prediction, historical insights from the target sector are retrieved
and incorporated into the current context together with their generation
metadata. We retrieve the most recent $K$ insights ordered by generation time.

In the original ExpeL setting, experiences are extracted from repeated task
trajectories with explicit success or failure outcomes. However, financial
prediction instances in FinEvolveBench exhibit limited repetition. Therefore,
we define each industry-level prediction as an experience episode and use the
realized market response as the delayed outcome signal. Insights are generated
separately for different prediction horizons to preserve horizon-specific
experience validity.

\subsection{ReasoningBank}

We adapt ReasoningBank to construct reasoning-level memories from historical
prediction trajectories. Each prediction episode is evaluated according to
whether the predicted direction agrees with the realized excess return. Based
on this outcome, successful and unsuccessful trajectories are processed using
different memory extraction prompts.

Each extracted memory contains a title, description, and content, and up to
three memory items are generated from a single trajectory. Since prediction
horizons represent different temporal objectives in FinEvolveBench, memories
are generated and stored separately for different horizons.

\subsection{Mem0}

We adapt Mem0 by organizing sector-level prediction records into experience memories that support semantic retrieval. Each memory contains a financial news summary, the model's predictions, and the historical memories referenced in the prediction. During prediction, the system uses the current news summary of the target sector as the query to retrieve semantically similar memories that were stored before the current prediction. The retrieved memories are then incorporated into the prediction context. After predictions are generated, the corresponding information is stored in the memory repository for subsequent retrieval, without waiting for market feedback.

To adapt Mem0 to FinEvolveBench, we shift the memory unit from user interactions to sector-level prediction episodes and use structured financial news summaries as retrieval queries. Sector filtering and temporal constraints ensure that retrieved memories belong to the target sector and are available before the current prediction. This implementation preserves Mem0's semantic memory storage and retrieval paradigm.

\subsection{MemRL}

We implement MemRL as a sector-level prediction memory mechanism that combines semantic retrieval with delayed utility feedback. Each memory contains a financial news summary, prediction results, the prediction date, and the historical memories referenced in the prediction, and maintains a value score $Q$. During prediction, the system first retrieves candidate memories based on vector similarity and then ranks them by their $Q$-values. The highest-ranked memories are incorporated into the current prediction context.

To adapt MemRL to the delayed feedback setting of FinEvolveBench, once the market outcome for a prediction horizon becomes available, the system computes a reward as the product of the prediction score and the sector's realized excess return relative to the market benchmark. This reward updates the $Q$-values of both the current prediction memory and the historical memories referenced in that prediction. Each memory maintains a single shared $Q$-value, which is updated separately by the rewards from the 10-, 20-, and 40-day predictions as their respective outcomes become available, thereby integrating feedback across horizons to estimate experience utility.

\section{Detailed Case Study}
\label{sec:case_study_appendix}

% We present a representative case where experience augmentation leads to
% negative transfer. This case corresponds to the transportation sector
% (\texttt{801170.SI}) on May 30, 2025. Given the available news context,
% agents predict the future sentiment direction over 10-, 20-, and 40-day
% horizons. As shown in Table~\ref{tab:case_prediction}, the experience-free
% pipeline correctly identifies a negative trend, while all experience-based
% methods produce positive predictions.

% \begin{table}[t]
% \centering
% \caption{
% Prediction results for the transportation case on May 30, 2025.
% The realized returns are provided as environment feedback after the
% corresponding prediction horizons.
% }
% \label{tab:case_prediction}
% \small
% \begin{tabular}{lccc}
% \toprule
% Method & 10d & 20d & 40d \\
% \midrule
% \textsc{Pipe} & -0.10 & -0.20 & -0.30 \\
% ExpeL & 0.70 & 0.40 & 0.20 \\
% Reflection & 0.70 & 0.50 & 0.30 \\
% Mem0 & 0.10 & 0.30 & 0.30 \\
% MemRL & 0.50 & 0.40 & 0.30 \\
% \midrule
% Realized Return & -2.37\% & -6.58\% & -10.75\% \\
% \bottomrule
% \end{tabular}
% \end{table}

\subsection{Representative News}
% 在表格~\ref{tab:case_news_raw}中，我们节选展示了transportation sector在2025年5月30日的四套具有代表性的新闻原始内容，展示的新闻中包含了两条26日的新闻，两条30日的新闻。这四条新闻分别展示了对行业走向的两种影响，正向的和负向的信号，这体现了我们环境的复杂性和任务的难度，也侧面体现了经验抽取和复用的难度。
Table~\ref{tab:case_news_raw} provides four representative news items
available before prediction for the transportation sector case.
The evidence includes both positive and negative signals across May 26 and
May 30, 2025: early indicators pointed to temporary demand expansion and
freight-rate increases, whereas later observations suggested potential demand
weakness. Such conflicting and evolving information reflects the uncertainty
of real-world decision environments, where agents must leverage historical
experience to interpret ambiguous signals and adapt their predictions.

\begin{table*}[h]
\centering
\caption{
Representative news evidence available before prediction.
Only information publicly available before May 30, 2025 is included.
}
\label{tab:case_news_raw}
\small
% \footnotesize
\renewcommand{\arraystretch}{0.5}
\begin{tabular}{p{2.2cm}p{11.5cm}}
\toprule
Date & News Content \\
\midrule

2025-05-26 &
\textbf{Surging Shipments Drive Freight Rates Sharply Higher, While Exporters Turn Cautious on New Orders.}
In the two weeks since China and the United States mutually reduced tariffs, previously backlogged orders from Chinese export enterprises have been shipped at an accelerated pace. According to multiple industry surveys, bookings on U.S.-bound shipping routes have surged continuously since May 12. Intensifying competition for limited vessel capacity has driven freight rates sharply higher in a nonlinear manner. As of now, shipping capacity for late May has been almost fully booked, with freight rates rising by more than 40\%.
The container shipping market is currently characterized by a tug-of-war between supply and demand. Freight forwarders expect freight rates to continue rising rather than falling over the next 90 days, while export enterprises anticipate that rates may decline once shipping capacity recovers in June. Amid the sharp increase in freight rates, shipping companies are urgently reallocating capacity to U.S.-bound routes. At the same time, some export enterprises have proactively slowed the pace of accepting new orders due to heightened uncertainty.
\\

2025-05-26 &
\textbf{U.S. Buyers Accelerate Stockpiling, Boosting Cargo Volumes at Shenzhen Port as Peak U.S.-Bound Shipping Season May Arrive Early.}
Following adjustments to China–U.S. tariff policies, U.S. buyers have accelerated inventory stockpiling, driving stronger demand for ocean freight on U.S.-bound routes. In response, many shipping companies have adjusted their capacity allocation, while port operations have become increasingly busy.
Shenzhen’s Yantian Port handles more than one-quarter of China’s total export cargo volume to the United States. During an on-site visit, Chen Zehao, duty manager of the Operations Services and Development Department at Yantian International Container Terminals, noted that daily container gate-in volumes have increased substantially compared with previous levels.
Traditionally, the peak shipping season runs from July to September, but this year it is expected to arrive earlier, in June and July. To prepare for the anticipated surge, the port has developed contingency plans covering cargo storage, empty-container repositioning, and yard operations, including optimizing yard utilization and increasing investment in equipment and operational resources.
\\

2025-05-30 &
\textbf{U.S. Goods Imports Fell 19.8\% in April, Recording the Largest Decline
on Record.}
According to the U.S. Census Bureau's ``Economic Indicators'' report, U.S.
goods imports decreased from \$344.470 billion in March to \$276.097 billion
in April, representing a decline of 19.8\%. The preliminary goods trade
deficit narrowed from \$162.3 billion in March to \$87.6 billion in April.
Thirty-four economists estimated the preliminary April goods trade deficit
within a range of \$110 billion to \$177.2 billion, with the median forecast
at \$143 billion.
\\

2025-05-30 &
\textbf{Industry Participants Expect a Significant Decline in May Cargo
Volume at the Port of Los Angeles.}
Due to recent fluctuations in U.S. tariff policies, port operations in the
United States have continued to be affected. Industry participants recently
stated that cargo volume at the Port of Los Angeles, the largest and busiest
container port in the United States, may decline significantly in May.
Gene Seroka, Executive Director of the Port of Los Angeles, stated on May 29
local time that although the traditional peak shipping season was approaching,
port activity was weaker than in previous years. Cargo volume at the port had
already declined by 30\% during the first and fourth weeks of May. He
expected that May cargo throughput at the Port of Los Angeles would decline
by double-digit percentages year over year. This could also affect employment
at the port, as fewer arriving vessels would reduce demand for cargo
unloading and transportation workers. Seroka further stated that ten
scheduled vessels originally expected to arrive at the Port of Los Angeles
in June had been canceled. The Port of Los Angeles is one of the busiest
ports in the United States and serves as the largest maritime gateway for
imports from China. Its major customers include large retailers such as
Walmart and suppliers of major automobile manufacturers such as Ford.
According to previously released data from the National Retail Federation,
U.S. imports in the second half of 2025 were expected to decline by at least
20\% year over year.
\\

\bottomrule
\end{tabular}
\end{table*}

\subsection{ExpeL}
\subsubsection{Experience retrieved by Expel}
\label{app:appendix_case_study_expel}

\begin{quote}
% \textbf{Source episode: Prediction 20250515, 10d}

The “rush shipment wave” triggered by China-U.S. tariff easing created a
strong and effective bullish driver for the transportation sector within
the 10-day horizon. Both the direction and magnitude of the prediction
were highly reliable. The prediction score of 0.7 and the actual excess
return change of 0.21 confirmed the immediate impact of trade policy
turning points on shipping and logistics.
\end{quote}

\paragraph{Analysis.}

This case reveals several limitations in how ExpeL constructs and reuses experiences. Although the retrieved experience correctly identifies a historical correlation between China-U.S. tariff easing and transportation gains, it fails to preserve the contextual conditions that determine when such a pattern remains valid. As a result, the experience is reused as a general bullish heuristic rather than a conditional strategy tied to specific market states. Furthermore, the abstraction process is primarily reinforced by successful outcomes, while insufficiently capturing counterexamples that characterize the failure boundaries of the learned pattern. The resulting experience also emphasizes the policy trigger itself while overlooking intermediate causal mechanisms, such as demand confirmation and inventory dynamics, that distinguish sustainable recovery from temporary demand front-loading. Finally, the experience representation remains too coarse-grained to capture horizon-dependent dynamics, where short-term shipment acceleration and medium-term demand normalization may lead to divergent outcomes.

\subsubsection{Prediction Reasoning}
\begin{quote}
This week, the transportation sector was affected by multiple intertwined factors. The core driver was the rush shipment wave on China-U.S. routes triggered by the mutual reduction of China-U.S. tariffs. Freight rates surged by more than 40\%, container capacity was sold out by the end of May, and shipping companies urgently adjusted capacity allocation. The short-term container shipping market was highly prosperous, directly benefiting shipping and port companies. However, U.S. commodity imports in April recorded a historic month-on-month decline of 19.8\%, and cargo volume at the Port of Los Angeles is expected to decline significantly in May, indicating that the rush shipment wave may have been a front-loaded effect, with subsequent demand potentially weakening and putting pressure on medium-term freight rates.

On the geopolitical side, the Gaza ceasefire proposal remained uncertain, and Houthi attacks on Israeli airports indicated continued uncertainty in the Red Sea situation. However, based on historical experience, geopolitical events have had limited and short-term suppression effects on the sector. On the policy side, the Ministry of Transport will introduce implementation guidelines for “Artificial Intelligence + Transportation,” and Shanghai has launched an international sustainable aviation fuel certification system, both representing long-term positive factors. The shipbuilding industry has full order books with production schedules extending to 2029, indicating sustained high industry prosperity.

Overall, the short-term rush shipment wave has pushed up freight rates and benefited the sector; however, in the medium term, attention should be paid to the risk of demand being brought forward and freight rate corrections caused by restored shipping capacity. In the long term, policy support and industrial upgrading provide continued support. Historical experience suggests that policy-driven catalysts should be viewed positively, but attention should also be paid to the correction risk after positive factors are realized.    
\end{quote}

\paragraph{Analysis.}
The prediction reasoning demonstrates that ExpeL is able to identify multiple
competing signals in the current environment, including the short-term
benefits from tariff-driven shipment acceleration and the medium-term risks
from demand front-loading. The reasoning also goes beyond simple news-level
association by considering intermediate mechanisms such as freight rate
dynamics, capacity adjustment, and inventory normalization.

However, the reasoning reveals a gap between evidence recognition and
experience adaptation. Although the model identifies several bearish signals,
including weakening U.S. import demand and potential freight rate correction,
the final reasoning still assigns substantial weight to the previously
observed pattern that tariff easing leads to transportation gains. This
indicates that the retrieved experience influences the decision process beyond
its original validity context.

More broadly, this case suggests that effective experience reuse requires not
only extracting relevant historical patterns but also dynamically evaluating
whether the underlying mechanism remains consistent with the current
environment. In low-repetition settings, where similar events may correspond
to different market states, successful reasoning depends on integrating past
experience with current evidence rather than transferring historical outcomes
directly.

\subsection{Reflexion}
\label{app:Reflexion}

\subsubsection{Experience retrieved by Reflexion}
\label{app:appendix_case_study_reflexion}

\begin{quote}
    \textbf{Summary of Prediction Performance}

**40-day forecast (+0.4; realized: +91.20 points): correct direction but severely underestimated magnitude.**  
At the prediction date, the market was simultaneously exposed to several major developments, including the Russia–U.S.–Ukraine agreement on Black Sea navigation security, continued U.S. airstrikes on Yemen with more than 40 strikes in a single day, and the issuance of the world’s first Operating Certificate (OC) for the low-altitude economy. The model assigned a moderately bullish score of +0.4, while the sector subsequently rose by +91.20 points over 40 days. The direction was correct, but the magnitude was substantially underestimated.
**Key belief update:** the freight-rate impact of the escalating Red Sea crisis— including Houthi attacks on U.S. aircraft carriers and nearly 70 U.S. airstrikes—continued to exceed expectations. At the same time, a major commercialization breakthrough in the low-altitude economy, marked by the world’s first operating certificate for passenger-carrying eVTOL services, created a reinforcing policy catalyst. The weights assigned to both “geopolitical escalation → higher freight rates” and “policy breakthroughs in the low-altitude economy” should therefore be increased.

**20-day forecast (-0.2; realized: +101.42 points): wrong direction and severely underestimated magnitude.**  
At the prediction date, negative developments were concentrated, including escalating India–Pakistan military tensions, Pakistan’s closure of its airspace and suspension of trade with India, the explosion at Iran’s Bandar Abbas port with the death toll rising to 70, and sharp declines in U.S. port cargo volumes, including a 44\% drop at the Port of Long Beach and a 33\% decline at the Port of Los Angeles. The model issued a mildly bearish score of -0.2, yet the sector subsequently rose sharply by +101.42 points over 20 days.
**Key belief update:** the negative effects of tariff shocks and geopolitical conflicts, including India–Pakistan tensions and the Bandar Abbas explosion, remained limited for the transportation sector. The market instead focused more heavily on positive policy catalysts, including the ten-department guideline on integrated development of transportation and energy, accelerated commercialization of the low-altitude economy, such as plans to enable urban air mobility in provincial capitals by 2027, and expanded departure-tax-refund policies designed to stimulate inbound consumption. This 20-day forecast was the worst-performing prediction among the three horizons in both directional accuracy and magnitude estimation.

**10-day forecast (+0.8; realized: +34.16 points): correct direction but underestimated magnitude.**  
At the prediction date, positive signals were concentrated, including the China–U.S. Geneva joint statement on economic and trade talks, under which 91\% of additional tariffs were removed and 24\% were suspended for 90 days; consecutive sharp gains in European-route container freight index futures, including a 16\% limit-up move and a cumulative weekly gain of 32\%; and a 277\% surge in China-to-U.S. container bookings. The model assigned a bullish score of +0.8, while the sector subsequently rose by +34.16 points over 10 days.
**Key validation:** the model still underweighted the mechanism of “China–U.S. tariff easing → order recovery → higher freight rates,” and underestimated the strength of the front-loading rush. Signals such as a nearly 300\% increase in China-to-U.S. container bookings, near-full utilization of U.S.-bound vessel capacity, and shipping-line price increases of USD 500–1,500 per container indicate that tariff easing generated a much stronger freight-rate effect than expected.

\textbf{Analysis of Logic Validity}

**40-day horizon:**  
The logic of “geopolitical escalation in the Red Sea / Strait of Hormuz → higher freight rates” continues to be validated. Continued Houthi attacks on U.S. aircraft carriers and more than 40 U.S. airstrikes on Yemen raised shipping risk premia, while the sector gained +91.20 points over 40 days. It is recommended that the 40-day model increase the weight of the “geopolitical escalation in key shipping corridors” factor from +1.8 to +2.0. The issuance of the world’s first low-altitude-economy operating certificate, with EHang’s EH216-S receiving approval for commercial operations, represents a new positive driver. A “low-altitude economy commercialization breakthrough” factor should therefore be added to the 40-day model with a weight of +0.5. Additional factors should also be monitored. CK Hutchison’s port transaction is subject to an antitrust review by China’s State Administration for Market Regulation and involves national critical-infrastructure security; a “port antitrust review” factor should therefore be incorporated with a weight of -0.1. Meanwhile, Trump’s threat to impose 25\%–50\% secondary tariffs on Russian oil could affect the global tanker market, so the “oil-shipping sanctions” factor should retain a weight of +0.2.

**20-day horizon:**  
The logic of “tariff shock → shrinking trade volumes → negative impact on transportation” continues to fail. Despite concentrated negative signals—including a 44\% decline in Long Beach cargo volumes, a 33\% decline at Los Angeles, Moody’s downgrade of the U.S. port outlook to negative, and the Bandar Abbas explosion in Iran that killed 70 people—the sector still rose by +101.42 points over 20 days.
The dominant driver remained the combination of “supportive policy resonance + long-term structural tailwinds,” including the ten-department guideline on transportation–energy integration, accelerated commercialization of the low-altitude economy, and expanded departure-tax-refund policies supporting inbound consumption.
It is recommended that the 20-day model reduce the negative weight of the “tariff shock” factor from -0.02 to -0.01, as its adverse effect remains limited but should not be removed entirely. The weight of the “free-trade zone / opening-up policy” factor should be increased from +3.0 to +3.5.
Additional signals should also be incorporated. JD.com plans to hire 100,000 full-time delivery riders over the next three months, all covered by social insurance and housing-fund benefits, suggesting labor expansion in the logistics sector; a “logistics labor expansion” factor should be added with a weight of +0.2. Construction has begun on key control projects of the China–Kyrgyzstan–Uzbekistan railway; a “Belt and Road infrastructure” factor should be added with a weight of +0.3. A large-scale power outage in Spain that also affected Portugal disrupted European supply chains; a “European supply-chain disruption” factor should be added with a weight of +0.1.

**10-day horizon:**  
The model predicted the correct direction (+0.8 versus a realized +34.16 points), but underestimated the magnitude. The China–U.S. Geneva joint statement—removing 91\% of additional tariffs and suspending another 24\% for 90 days—became the dominant positive catalyst. Consecutive sharp gains in European-route container freight futures, including a 16\% limit-up move and a 32\% weekly gain, a 277\% increase in China-to-U.S. container bookings, near-full utilization of U.S.-bound capacity, and freight-rate increases of USD 500–1,500 per container all validated the mechanism of “tariff easing → order recovery → higher freight rates.”
It is recommended that the 10-day model increase the weight of the “China–U.S. tariff easing” factor from +1.5 to +2.0, given its strong validation in this episode. The weight of the “container freight index surge / higher freight rates” factor should also be increased from +1.8 to +2.0, as the front-loading effect continued to be confirmed.
Negative signals such as shrinking U.S. port volumes—including zero cargo vessels at the Port of Seattle and a 33\% decline at the Port of Los Angeles—had only a limited impact on the A-share transportation sector, so the “U.S. port contraction” factor should remain at -0.1.
Additional factors should be incorporated. Beginning May 15, the Suez Canal Authority offered a 15\% transit-fee discount to large container ships in an effort to encourage their return to the canal, implying expectations of improving Red Sea security. A “Suez Canal discount” factor should be added with a weight of -0.2, reflecting a mildly negative effect on freight rates. Qatar Airways placed an order for at least 160 Boeing wide-body aircraft valued at USD 200 billion, supporting the aviation manufacturing industry; an “aircraft manufacturing orders” factor should be added with a weight of +0.2. CMA CGM also announced that rates for large containers on U.S.-bound routes would rise to USD 10,000 after mid-June, reinforcing expectations of higher freight rates; a “freight-rate increase expectations” factor should be added with a weight of +0.3.

\textbf{Interpretation of Spread Signals}

The 40-day realized change of +91.20 points was directionally consistent with the predicted score of +0.4 but substantially larger in magnitude. The signal can be interpreted as follows: continued Houthi attacks on U.S. aircraft carriers strengthened expectations of higher freight rates through the escalation of the Red Sea crisis; the issuance of the world’s first low-altitude-economy OC for EHang’s EH216-S marked the beginning of commercial passenger operations; and the action plan for connectivity improvements in the inland-waterway transport system provided additional policy support. Together, these factors drove a substantial +91.20-point increase in the sector.
The model correctly identified the direction, but still underestimated the combined effect of geopolitical conflict on freight rates and supportive policy catalysts. At the same time, easing signals—including the Russia–U.S. agreement on Black Sea navigation security and U.S. officials’ statements that key Houthi leaders had been successfully targeted—may constrain further upside in freight rates.

The 20-day realized change of +101.42 points was completely opposite to the predicted score of -0.2 and was severely underestimated. Despite concentrated negative developments—including escalating India–Pakistan military tensions, Pakistan’s closure of its airspace and suspension of trade with India, the Bandar Abbas explosion in Iran with the death toll rising to 70, and sharp declines in U.S. port cargo volumes—the sector rose rather than fell.
The model therefore got the direction wrong. The negative transmission of tariff shocks and geopolitical conflicts to the transportation sector remained limited, while supportive policies—including the transportation–energy integration guideline, accelerated commercialization of the low-altitude economy, and expanded departure-tax-refund policies—became dominant. The market appeared to interpret tariff-related pressure more as a trigger for policy offset than as evidence of a collapse in industry demand.

The 10-day realized change of +34.16 points was directionally consistent with the predicted score of +0.8 but larger in magnitude. The China–U.S. Geneva joint statement, consecutive sharp increases in European-route container freight futures, a 277\% surge in China-to-U.S. container bookings, near-full utilization of U.S.-bound shipping capacity, and freight-rate increases of USD 500–1,500 per container collectively drove the sector higher.
The model correctly captured the direction. The mechanism of “tariff easing → order recovery → higher freight rates” was validated, but the magnitude was underestimated because the strength of the front-loading rush exceeded expectations.

\textbf{Lessons and Points to Monitor}

1. **The 40-day weight of “geopolitical escalation in the Red Sea / Strait of Hormuz” should be increased to +2.0.**  
This episode confirms that continued Houthi attacks on U.S. aircraft carriers and more than 40 U.S. airstrikes on Yemen raised shipping risk premia, while the sector gained +91.20 points over 40 days. The weight of the “geopolitical escalation” factor should therefore be increased from +1.8 to +2.0. Although the Russia–U.S. Black Sea navigation agreement and U.S. claims of successful strikes against key Houthi leaders may constrain further freight-rate increases, the Red Sea crisis remains the dominant source of support for freight rates.

2. **The 20-day weight of “free-trade zone / opening-up policy” should be increased to +3.5.**  
Despite escalating India–Pakistan tensions, the Bandar Abbas explosion, and sharp declines in U.S. port cargo volumes, the sector rose by +101.42 points over 20 days, only slightly below the previous episode’s +106.03 points and still at a high level. The weight of the “free-trade zone / opening-up policy” factor should therefore be increased from +3.0 to +3.5. When tariff-related headwinds and supportive policies occur simultaneously, the market continues to place greater weight on policy-offset effects, while the negative transmission of tariffs to the transportation sector remains limited.

3. **The 20-day weight of the “tariff shock” factor should be reduced to -0.01.**  
Although tariff-related negative signals were concentrated—including a 44\% decline in Long Beach cargo volumes, a 33\% decline at Los Angeles, and Moody’s downgrade of the U.S. port outlook—the sector still rose by +101.42 points. The weight of the “tariff shock” factor should therefore be reduced from -0.02 to -0.01. Its negative impact on the transportation sector remains limited, while the market focuses more heavily on policy-offset effects.

4. **The 10-day weight of “China–U.S. tariff easing” should be increased to +2.0.**  
Following the China–U.S. Geneva joint statement, European-route container freight futures surged, China-to-U.S. container bookings increased by 277\%, U.S.-bound capacity approached full utilization, and shipping companies raised rates by USD 500–1,500 per container. The sector subsequently gained +34.16 points over 10 days. The weight of “China–U.S. tariff easing” should therefore be increased from +1.5 to +2.0. The front-loading rush triggered by tariff easing had a particularly strong effect on freight rates, and the model continued to underweight this mechanism.

5. **The “front-loading rush” should be introduced as an independent factor.**  
Following the mutual tariff reductions between China and the United States, China-to-U.S. container bookings surged by 277\%, U.S.-bound shipping capacity approached full utilization, freight rates increased by USD 500–1,500 per container, and CMA CGM announced that rates for large containers on U.S.-bound routes would rise to USD 10,000 after mid-June. A new “front-loading rush” factor with a weight of +1.5 is recommended, reflecting its strong short-term effect on freight rates. Whenever tariff-easing signals emerge, the model should jointly assess the strength of front-loading demand and its impact on freight rates.

6. **The weight of the “low-altitude economy” factor should be increased to +0.8.**  
The issuance of the world’s first operating certificate for a civil unmanned passenger aircraft, covering EHang’s EH216-S, marked the beginning of commercial passenger operations in the low-altitude economy. In addition, the second Global Low-Altitude Economy Forum proposed enabling urban air mobility in provincial capitals by 2027 and gradually delegating airspace management authority to prefecture-level cities, while commercial low-altitude logistics continued to accelerate. The weight of the “low-altitude economy” factor should therefore be increased from +0.5 to +0.8. Continued policy support for low-altitude logistics could provide a long-term structural tailwind for the transportation sector.

7. **The mechanism through which geopolitical conflict affects the transportation sector should be further decomposed.**  
This episode shows that escalating India–Pakistan tensions, including Pakistan’s closure of its airspace and suspension of trade with India, created short-term pressure on the aviation segment, yet the sector still rose by +101.42 points over 20 days. The impact of geopolitical conflict on transportation should therefore not be modeled as a single factor. The model should distinguish between “geopolitical conflict affecting shipping supply,” such as the Red Sea crisis, which may raise freight rates and benefit shipping companies, and “geopolitical conflict disrupting aviation operations,” such as airspace closures, which is negative but appears to carry relatively limited weight.

Particular attention should be paid to the Bandar Abbas explosion in Iran. As one of Iran’s most important ports, handling more than 55\% of the country’s non-oil imports and exports, disruptions there could affect global shipping chains, although the impact on the A-share transportation sector in this episode remained limited.

8. **The “Suez Canal discount” signal should be incorporated into the factor system.**  
Beginning May 15, Egypt’s Suez Canal Authority offered a 15\% transit-fee discount to large container vessels in an attempt to encourage shipping companies to return to the canal, reflecting expectations of an improvement in Red Sea security. A “Suez Canal discount” factor with a weight of -0.2 is recommended. If security conditions in the Red Sea continue to improve, freight rates may face downward pressure, although the supportive effect of the ongoing Red Sea crisis remains dominant.

9. **An “aircraft manufacturing orders” factor should be incorporated into the model.**  
Qatar Airways placed orders for at least 160 Boeing wide-body aircraft valued at approximately USD 200 billion, while Etihad Airways made procurement commitments worth USD 14.5 billion to Boeing and General Electric. These large aviation orders represent positive catalysts for the aircraft manufacturing supply chain. An “aircraft manufacturing orders” factor with a weight of +0.2 is recommended, while recognizing that its transmission to the A-share transportation sector is likely to remain limited.
\end{quote}

\paragraph{Analysis.}
This case reveals several limitations in how Reflexion forms and reuses reflective experiences. Although the retrieved reflection correctly identifies the historical association between tariff easing and transportation gains, it fails to capture the contextual conditions under which such a relationship remains valid. The reflection mainly summarizes successful outcomes from previous episodes and reinforces the heuristic that tariff-related improvements indicate a bullish transportation outlook, while insufficiently analyzing the boundary conditions that determine when this pattern may fail.

Furthermore, the reflection emphasizes the observable policy trigger while overlooking intermediate causal mechanisms, such as front-loading demand, inventory adjustment, and subsequent demand normalization. Although tariff easing generated short-term transportation gains through accelerated shipment demand, later weakening signals suggested that part of the improvement was temporary rather than a sustained recovery. Without deeper causal analysis, Reflexion transfers a context-dependent pattern into future scenarios as a general strategy.

The retrieved reflection is also limited by its coarse representation granularity. Transportation markets exhibit horizon-dependent dynamics, where the same factor may produce different outcomes over short and medium horizons. However, the reflection encodes tariff easing as a directional bullish signal rather than representing the underlying temporal transition from short-term shipment acceleration to potential medium-term normalization.

Finally, Reflexion lacks explicit mechanisms for revising previously formed reflections when new evidence conflicts with earlier conclusions. Once the bullish reflection was established, negative signals such as weakening imports and port activity were insufficiently incorporated to invalidate or weaken the previous reasoning pattern, limiting its adaptability under non-stationary environments.

\subsubsection{Prediction Reasoning}

\begin{quote}
The core driver of the transportation sector this week was the “front-loading rush” triggered by the easing of China–U.S. tariffs. Bookings on U.S.-bound routes surged, vessel capacity was largely sold out, freight rates jumped by more than 40\%, container freight index futures rose sharply, and shipping companies raised their rates, directly benefiting the short-term performance of container shipping and port operators.

On the geopolitical front, developments surrounding the Gaza ceasefire proposal remained volatile, with Hamas expressing acceptance while the United States and Israel denied agreement, followed by Israel indicating acceptance while Hamas remained dissatisfied. Meanwhile, the Houthis continued attacks on Israeli airports. As a result, the Red Sea crisis has not been fundamentally resolved, continuing to provide support for freight rates.

On the policy side, the Ministry of Transport is expected to issue implementation guidelines for “AI + Transportation,” Shanghai has launched an international certification system for sustainable marine fuels, and the ASEAN–China–GCC Summit explored cooperation in local-currency settlement and cross-border payments. These developments constitute medium- to long-term positive catalysts.

However, downside risks should not be overlooked. U.S. goods imports fell by a record 19.8\% month on month in April, while cargo volumes at the Port of Los Angeles are expected to decline by a double-digit percentage in May, indicating that the negative impact of tariff-related disruptions on trade volumes is beginning to emerge. In addition, some export-oriented firms have slowed the pace of accepting new orders due to uncertainty, which may constrain further upside in freight rates.

Overall, in the short term (10–20 days), the front-loading rush and geopolitical risk premium are likely to remain the dominant drivers, implying a moderately bullish sentiment. Over the medium term (40 days), greater attention should be paid to the lagged impact of weakening trade volumes, suggesting a neutral-to-moderately bullish outlook.
\end{quote}

\paragraph{Analysis.} 
The prediction reasoning shows that Reflexion can identify both positive and
negative signals in the current environment, including the short-term impact
of tariff-driven shipment acceleration and the medium-term risk from demand
normalization. However, the retrieved reflection still biases the reasoning
toward the previously observed pattern that tariff easing leads to
transportation gains. Although the model recognizes weakening trade signals,
it does not sufficiently analyze the conditions under which the historical
pattern becomes invalid.

This failure indicates that the reflection mainly captures the observed
relationship between events and outcomes, while lacking deeper causal
decomposition of the underlying mechanism. In this case, the reflection
emphasizes tariff easing as the key driver but fails to distinguish temporary
front-loading demand from sustainable recovery. As a result, short-term
success patterns are transferred to a broader market context where their
validity is uncertain.

More broadly, this case suggests that effective reflective experience reuse
requires reflections to preserve not only past conclusions but also the
causal boundaries and failure conditions of those conclusions. Without such
contextualized reflection, Reflexion may recognize conflicting evidence yet
remain anchored to outdated reasoning patterns in non-stationary
environments.

\subsection{Mem0}
\label{app:Mem0}

\subsubsection{Experience retrieved by Mem0}
\begin{quote}
\textbf{Memory 01}

*Similarity\_score: 0.9353379

*Prediction Date: 20250529

*Sector: Transportation (801170.SI)

*News Window: 2025-05-25 ~ 2025-05-29

*High-Importance Market-Wide News:

[2025-05-29 23:55:35+08:00] id=41014 src=sina importance=10 topic=Geopolitics, Policy Planning, Cross-Market Linkages industries=All Sectors
  title: [Trump Administration Asks Appeals Court to Pause Tariff Ruling] The Trump administration has asked an appeals court to pause the tariff ruling. The U.S. government said it may ask the Supreme Court on Friday to keep the tariffs in place. Note: On May 28 local time, the U.S. Court of International Trade...
  content: [Trump Administration Asks Appeals Court to Pause Tariff Ruling] The Trump administration has asked an appeals court to pause the tariff ruling. The U.S. government said it may ask the Supreme Court on Friday to keep the tariffs in place. Note: On May 28 local time, the U.S. Court of International Trade blocked the tariff policy announced by U.S. President Donald Trump on April 2, the so-called "Liberation Day," and ruled that Trump had exceeded his authority by imposing broad tariffs on countries that export more goods to the United States than they import from it.
  importance\_reason: The event concerns a court decision blocking the U.S. president's tariff policy, followed by the Trump administration seeking intervention from an appeals court and potentially the Supreme Court. It directly affects the global trade order and market expectations, with implications across global markets, making it a highly significant geopolitical and trade-policy event.

...

*High-Importance Sector-Specific News:

[2025-05-29 23:52:53+08:00] id=41025 src=jinrongjie importance=10 topic=Geopolitics, Cross-Market Linkages industries=Petroleum \& Petrochemicals, Transportation
  title: Israeli Prime Minister Accepts U.S.-Proposed Gaza Ceasefire Plan
  content: Israeli media reported on May 29 local time that Israeli Prime Minister Benjamin Netanyahu had accepted a U.S.-proposed ceasefire plan for the Gaza Strip.
  importance\_reason: This news concerns the Israeli prime minister's acceptance of a U.S.-proposed ceasefire plan for the Gaza Strip and therefore constitutes a major geopolitical event. A ceasefire in the Gaza conflict could directly affect stability in the Middle East and have immediate and broad implications for global energy prices, safe-haven assets such as gold and the U.S. dollar, and shipping security through the Suez Canal. It is therefore a highly significant event with potential impacts across global markets.

...

*Prediction:
\{
"confidence\_score": 0.75,
"key\_factor": "Easing Middle East tensions and tariff pressures",
"logic\_chain": "The transportation sector this week was influenced by multiple overlapping geopolitical developments, with the core logic centered on the expected easing of tensions in the Middle East and a shift in global trade policy. First, the Israeli prime minister accepted a Gaza ceasefire proposal, following Hamas's earlier agreement to a ceasefire plan. This strengthened expectations of de-escalation in the Middle East and directly benefited shipping security, as risks along the Red Sea and Suez Canal routes were expected to decline. Lower security risks could reduce marine insurance costs and rerouting expenses, providing both sentiment and fundamental support to container shipping and oil transportation. Second, the U.S. Court of International Trade ruled that Trump's broad tariff measures exceeded presidential authority and blocked their implementation. Although the administration appealed the ruling, the judicial decision signaled a potential turning point, strengthening expectations of an easing in the global trade conflict and improving the outlook for cross-border logistics and port throughput. In addition, U.S.-bound shipping experienced a front-loading rush following mutual tariff reductions, with freight rates rising by more than 40\%, providing a short-term boost to shipping companies' profitability. However, uncertainties remain, as Israel denied that a ceasefire agreement had been reached and Houthi missile attacks continued. Meanwhile, the delisting of *ST Jinzhou Port may exert localized pressure on sentiment toward the port sector. Overall, sentiment is moderately positive in the short term (1--5 days). Over the medium term (10--20 days), developments in ceasefire implementation and the tariff appeal process should be closely monitored. If the current trends continue, the positive effects are expected to persist."
\}
\end{quote}

\paragraph{Analysis.}

This retrieved memory reveals a fundamental limitation of trajectory-based
experience reuse: the retrieved item represents a past prediction rather than a
validated experience. Although the memory is highly similar to the current
case in terms of sector and event context, it only records how the agent
interpreted the information at that time, without indicating whether this
interpretation was subsequently confirmed or remained reliable under similar
conditions.

From the perspective of experience formation, the memory directly associates
tariff relief and geopolitical developments with a bullish transportation
outlook, but does not contain outcome-based calibration of this pattern.
Specifically, there is no evidence that this relationship was repeatedly
validated across multiple episodes, nor any indication of how its reliability
should change after successful or failed predictions. As a result, an
unverified trajectory can be reused in the same manner as a confirmed
experience.

The retrieved representation also provides limited support for transfer
across environments. It compresses heterogeneous drivers, including freight
rates, trade demand, and geopolitical risks, into a single sector-level
prediction trajectory, without preserving the conditions under which these
factors produced the historical outcome. Therefore, the memory captures
surface-level similarity while lacking information required to judge whether
the underlying pattern remains applicable.

This case demonstrates that episodic memory retrieval requires more than
finding similar historical trajectories. Reliable experience construction
requires associating memories with verified outcomes, reliability estimates,
and contextual validity.

\subsubsection{Prediction Reasoning}
\begin{quote}
The transportation sector was influenced this week by a combination of geopolitical developments and trade-policy events. The core logic is as follows. First, a U.S. court ruled that Trump's broad tariff measures exceeded presidential authority and blocked their implementation, while the Trump administration subsequently asked an appeals court to stay the ruling. This has created substantial uncertainty around the future path of tariff policy, weighing on global trade expectations in the short term; however, if the tariffs are ultimately removed, the medium- to long-term outlook for cross-border logistics and shipping demand would improve.

Second, geopolitical tensions in the Middle East remain volatile. The ceasefire negotiations between Hamas and Israel have gone through repeated reversals, while Houthi forces have continued attacks on Israeli airports. As a result, security risks in Red Sea shipping remain elevated, supporting higher shipping risk premia and freight rates.

Third, U.S. goods imports fell by 19.8\% month on month in April, indicating a sharp contraction in trade activity under the impact of tariffs and creating short-term headwinds for transportation demand, particularly for ports and shipping.

Fourth, freight rates on U.S.-bound routes surged by more than 40\% as shippers rushed to move cargo following the mutual tariff reductions, although demand may weaken afterward.

Fifth, China's shipbuilding order books are reportedly filled through 2029, indicating a sustained upcycle in the shipbuilding industry and providing support for transportation-related segments connected to vessel manufacturing.

Overall, short-term market sentiment remains cautious due to tariff uncertainty and geopolitical disruptions. Over the medium to long term, however, the sector could recover if the trade environment improves and a ceasefire in the Middle East is successfully implemented.
\end{quote}

\paragraph{Analysis.}
The prediction reasoning recognizes several signals that contradict the
retrieved memory, including weakening U.S. import demand, uncertainty in trade
recovery, and the possibility that recent shipment acceleration reflects
temporary front-loading rather than sustainable growth. However, these signals
are incorporated mainly as additional considerations rather than as evidence
to reassess the applicability of the retrieved experience.

The main limitation lies in the causal interpretation of the retrieved pattern.
The reasoning continues to rely on the historical association between tariff
easing and transportation gains, but does not explicitly distinguish the
mechanisms that generate short-term and medium-term outcomes. In particular,
it does not separate temporary freight-rate increases caused by front-loading
demand from sustained recovery driven by fundamental trade improvement.
Consequently, contradictory evidence weakens the prediction confidence but
does not fundamentally revise the underlying experience.

Furthermore, the reasoning represents transportation performance as a unified
sector-level outcome, while the retrieved pattern involves multiple mechanisms
with different temporal dynamics. Freight rates, trade volumes, and policy
support may contribute differently across prediction horizons, yet the
reasoning does not establish how these factors should be weighted when market
conditions change.

This case highlights that effective experience reuse requires not only
retrieving relevant past patterns but also evaluating whether their causal
mechanisms remain consistent with the current environment. Without such
context-aware reasoning, agents may acknowledge conflicting evidence while
still being anchored by outdated experiences.

\subsection{Memrl}
\label{app:Memrl}

\subsubsection{Experience retrieved by MemRL}
\begin{quote}

\textbf{Memory 01}

*Similarity: 1.0

*Memrl\_score: 0.85

*Prediction Date: 20250529

*Sector: Transportation (801170.SI)

*News Window:2025-05-25 \~ 2025-05-29

*High-Importance Market-Wide News:

[2025-05-29 23:55:35+08:00] id=41014 src=sina importance=10 topic=Geopolitics, Policy Planning, Cross-Market Linkages industries=All Sectors
  title: [Trump Administration Requests Appeals Court to Stay Tariff Ruling] The Trump administration has asked an appeals court to stay the tariff ruling. The U.S. government said it may ask the Supreme Court on Friday to keep the tariffs in place. Note: On May 28 local time, the U.S. Court of International Trade...
  content: [Trump Administration Requests Appeals Court to Stay Tariff Ruling] The Trump administration has asked an appeals court to stay the tariff ruling. The U.S. government said it may ask the Supreme Court on Friday to keep the tariffs in place. Note: On May 28 local time, the U.S. Court of International Trade blocked the tariff policy announced by U.S. President Donald Trump on April 2, known as ``Liberation Day,'' and ruled that Trump had exceeded his authority by imposing broad tariffs on countries that export more goods to the United States than they import from it.
  importance\_reason: The event concerns a court ruling that blocked the U.S. president's tariff policy, followed by the Trump administration seeking intervention from an appeals court and potentially the Supreme Court. It directly affects the global trade order and market expectations, with implications across global markets, making it a highly significant geopolitical and trade-policy event.

...

*High-Importance Sector-Specific News:

[2025-05-29 23:52:53+08:00] id=41025 src=jinrongjie importance=10 topic=Geopolitics, Cross-Market Linkages industries=Petroleum \& Petrochemicals, Transportation
  title: Israeli Prime Minister Accepts U.S.-Proposed Gaza Ceasefire Plan
  content: Israeli media reported on May 29 local time that Israeli Prime Minister Benjamin Netanyahu had accepted a U.S.-proposed ceasefire plan for the Gaza Strip.
  importance\_reason: This news concerns the Israeli prime minister's acceptance of a U.S.-proposed ceasefire plan for the Gaza Strip and therefore constitutes a major geopolitical event. A ceasefire in the Gaza conflict could directly affect stability in the Middle East and have immediate and broad implications for global energy prices, safe-haven assets such as gold and the U.S. dollar, and shipping security through the Suez Canal. It is therefore a highly significant event with potential impacts across global markets.

...

*Prediction

From the dual perspectives of geopolitics and macroeconomic policy, the core drivers of the transportation sector this week were the U.S. tariff policy being ruled to exceed presidential authority and temporarily blocked by the court, representing a major positive development, together with a breakthrough in Middle East ceasefire negotiations, as the Israeli prime minister accepted a ceasefire proposal, and improving industry fundamentals, including surging freight rates on U.S.-bound routes and an earlier-than-usual shipping peak.

Specifically: 1) The U.S. Court of International Trade ruled that Trump's broad tariff measures exceeded presidential authority and blocked their implementation. Although the Trump administration requested that an appeals court stay the ruling, the judicial decision had already determined that the president had exceeded his authority. This directly reversed the previous pessimistic expectations of an escalating global trade war and provided strong short-term support for shipping and cross-border logistics. Historical experience also suggests that Trump's trade strategy tends to follow a recurring ``threat--implementation--panic--reversal'' pattern, implying that the sector could recover over the medium term if negotiations lead to another policy reversal.

2) In the Middle East, the Israeli prime minister accepted a U.S.-proposed Gaza ceasefire plan on May 29, following Hamas's earlier acceptance of a ceasefire proposal on May 26. Although the process involved repeated setbacks, including Israeli denials and Houthi attacks, Israel's eventual acceptance of the ceasefire proposal represents a major positive development. It directly alleviates concerns over shipping security in the Red Sea and Suez Canal, lowers expected marine insurance costs, and provides strong support for the shipping segment.

3) In terms of industry fundamentals, freight rates on U.S.-bound routes surged by more than 40\%, vessel capacity was largely sold out, and the peak shipping season was brought forward to June--July, directly improving the earnings outlook for shipping companies.

4) Additional policy support, including the European Union's removal of economic sanctions on Syria and efforts by the ASEAN--China--GCC Summit to advance free-trade cooperation, further supports trade and logistics demand.

Overall, the combination of a reversal in U.S. tariff policy, progress toward a Middle East ceasefire, and strong industry fundamentals generates a strongly bullish signal for the transportation sector. However, uncertainty remains regarding the Trump administration's potential appeal to the U.S. Supreme Court and the implementation risks surrounding the Middle East ceasefire.
\end{quote}

\paragraph{Analysis.}
This retrieved experience reveals a limitation of outcome-based memory scoring:
a high utility score does not necessarily indicate that the underlying
mechanism has been reliably identified. Although the memory receives a high
MemRL score (0.85), it represents a bullish prediction from a single episode
rather than a validated rule describing when tariff easing and geopolitical
changes consistently lead to transportation gains.

The main limitation lies in how outcome feedback is converted into experience
quality. The score indicates that this memory has accumulated value, but the
stored experience does not reveal whether the value comes from repeated
successful outcomes, similarity to a specific market regime, or reinforcement
of a particular prediction trajectory. Consequently, MemRL may assign high
utility to experiences whose conclusions are correct only under limited
conditions.

Moreover, the retrieved experience compresses multiple mechanisms, including
tariff policy, freight-rate changes, and geopolitical developments, into a
single bullish sector-level pattern. Such a representation does not preserve
which component contributed to the observed outcome or under what conditions
the relationship breaks down. Therefore, high retrieval priority alone cannot
guarantee that the experience remains transferable to a changing environment.

\subsubsection{Prediction Reasoning}
\begin{quote}
From the perspectives of geopolitics and macroeconomic policy, the main drivers of the transportation sector this week are the court ruling that blocked the U.S. tariff policy as exceeding presidential authority, a breakthrough in Middle East ceasefire negotiations following the Israeli prime minister’s acceptance of a ceasefire proposal, and favorable industry developments, including a surge in freight rates on U.S.-bound routes and an earlier shipping peak.

Specifically, (1) the U.S. Court of International Trade ruled that Trump’s broad tariffs exceeded presidential authority and blocked their implementation. Although the administration asked an appeals court to stay the ruling, the decision reversed earlier pessimistic expectations of an escalating global trade war, providing strong short-term support for shipping and cross-border logistics. The reasoning also invokes a historical “threat–implementation–panic–reversal” pattern in Trump’s trade policy, suggesting that negotiations could support a medium-term recovery. (2) In the Middle East, the Israeli prime minister reportedly accepted a U.S.-proposed Gaza ceasefire plan on May 29, following Hamas’s acceptance of a proposal on May 26. Despite intervening setbacks, including Houthi attacks on an Israeli airport, the prospect of a ceasefire eased concerns about shipping security in the Red Sea and Suez Canal and could lower marine insurance costs, benefiting shipping companies. (3) On industry fundamentals, a rush to ship goods following reciprocal tariff reductions drove freight rates on U.S.-bound routes up by more than 40\%; vessel capacity sold out, and shipping companies urgently reallocated capacity. However, U.S. goods imports fell by a record 19.8\% in April, while cargo volumes at the Port of Los Angeles declined sharply. These figures suggest that tariff uncertainty has already affected trade volumes and raise concerns about medium-term demand. (4) Elsewhere, U.S. export controls on engines for the C919 weigh on aircraft manufacturing, though their effect on the broader transportation sector is relatively limited. The delisting of *ST Jinzhou Port also puts localized pressure on port-sector sentiment.

Overall, the apparent tariff reversal and progress toward a Middle East ceasefire reinforce a strongly optimistic short-term outlook (1–5 days). Over the medium term (10–20 days), sentiment is expected to moderate as the outcome of the tariff appeal, implementation of the ceasefire, and trade data provide further evidence on demand.
\end{quote}

\paragraph{Analysis.}
This prediction demonstrates that MemRL can retrieve and incorporate
historical experience while also recognizing several conflicting signals,
including weakening U.S. import demand, declining port activity, and the risk
that shipment acceleration reflects temporary front-loading. However, these
signals are treated as secondary risks rather than evidence that should
reassess the retrieved experience.

The key failure lies in the interaction between retrieved experience and
current evidence. Although MemRL assigns utility to historical experiences, the
reasoning does not clarify how the retrieved score should be adjusted when the
current environment violates the assumptions behind the memory. In this case,
the historical pattern of tariff easing leading to transportation gains is
accepted as a dominant prior, while the evidence that the mechanism may reflect
temporary demand acceleration is not used to revise that belief.

Furthermore, the final prediction aggregates heterogeneous effects across
different horizons into a single sector-level judgment. Short-term freight
rate increases may benefit shipping operators, whereas weakening trade demand
may indicate pressure on subsequent transportation activity. Without explicitly
separating these mechanisms, the retrieved experience continues to influence
the decision despite unresolved conflicts with current evidence.

\end{document}